\documentclass[10pt,twocolumn,letterpaper]{article}

\usepackage[pagenumbers]{iccv} % To force page numbers, e.g. for an arXiv version

\usepackage{graphicx}%
\usepackage{multirow}%
\usepackage{amsmath,amssymb,amsfonts}%
\usepackage{mathrsfs}%
\usepackage[title]{appendix}%
\usepackage{textcomp}%
\usepackage{manyfoot}%
\usepackage{booktabs}%
\usepackage{enumitem}
\usepackage{algorithm}%
\usepackage{algorithmicx}%
\usepackage{algpseudocode}%
\usepackage{listings}%
\usepackage{booktabs}%
\usepackage{adjustbox}%
\usepackage{tabularx}%
\usepackage{diagbox}%
\usepackage{caption}%
\usepackage{subcaption, soul}%
\usepackage{booktabs}
\usepackage{colortbl}
\usepackage{algorithm}
\usepackage{svg}

\usepackage{listings}

\usepackage{acronym}
\newacro{gan}[GAN]{generative adversarial network}
\newacro{vae}[VAE]{variational autoencoder}
\newacro{cnn}[CNN]{convolutional neural network}
\newacro{dire}[DIRE]{diffusion reconstruction error}
\newacro{mlp}[MLP]{multi-layer perceptron}
\newacro{clip}[CLIP]{contrastive language image pre-training}
\newacro{vlm}[VLM]{vision-language model}
\newacro{vit}[ViT]{vision transformer}
\newacro{nlp}[NLP]{natural language processing}
\newacro{cv}[CV]{computer vision}
\newacro{blip}[BLIP]{bootstrapping language image pre-training}
\newacro{vqa}[VQA]{visual question answering}
\newacro{lora}[LoRA]{low-rank adaptation}
\newacro{peft}[PEFT]{parameter-efficient fine-tuning}
\newacro{gpt}[GPT]{generative pre-trained transformer}
\newacro{q-former}[Q-Former]{querying transformer}
\newacro{sedid}[SeDID]{stepwise error for diffusion-generated image detection}
\newacro{sd}[SD]{stable diffusion}
\newacro{lsun}[LSUN]{large-scale scene understanding}
\newacro{fc}[FC]{fully-connected}
\newacro{deit}[DeiT]{data-efficient image transformers}
\newacro{ldm}[LDM]{latent diffusion model}
\newacro{adm}[ADM]{ablated diffusion model}
\newacro{ddpm}[DDPM]{denoising diffusion probabilistic models}
\newacro{iddpm}[IDDPM]{improved denoising diffusion probabilistic models}
\newacro{pndm}[PNDM]{pseudo numerical methods for diffusion models on manifolds}
\newacro{srm}[SRM]{spatial rich model}
\newacro{lasted}[LASTED]{language-guided synthesis detection}
\newacro{rf}[RF]{random forest}
\newacro{dm}[DM]{diffusion model}
\newacro{ddim}[DDIM]{denoising diffusion implicit models}
\newacro{multilid}[multiLID]{multi local intrinsic dimensionality}
\newacro{ifdl}[IFDL]{image forgery detection and localization} 
\newacro{svm}[SVM]{support vector machine} 
\newacro{ai}[AI]{artificial intelligence} 

\newacro{amsff}[AMSFF]{attention-based multi-scale feature fusion} 
\newacro{psm}[PSM]{patch selection module}
\newacro{llm}[LLM]{large language model}
\newacro{clip}[CLIP]{Contrastive Language-Image Pretraining}
\newacro{c2p}[C2P]{category common prompt}
\newacro{sid}[SID]{synthetic image detection}
\newacro{fosid}[FOSID]{Fact-checked Online Synthetic Image Dataset}
\newacro{rasid}[RASID]{Retrieval-Assisted Synthetic Image Detection}
\newacro{gff}[GFF]{Guided and Fused Frozen CLIP-ViT}
\newacro{fuseformer}[FuseFormer]{Multi-Stage Fusion Module}
\newacro{dfgm}[DFGM]{Deepfake-Specific Feature Guidance Module}
\newacro{ffaa}[FFAA]{Face Forgery Analysis Assistant}
\newacro{mllm}[MLLM]{ Multimodal Large Language Model}
\newacro{mids}[MIDS]{Multi-answer Intelligent Decision System}
\newacro{ssm}[SSM]{state space model}
\newacro{einfft}[EinFFT]{Einstein FFT}
\newacro{fft}[FFT]{Fast Fourier Transform}
\newacro{vim}[ViM]{Vision Mamba}
\newacro{ms2d}[MS2D]{Multi-Scale 2D}
\newacro{convffn}[ConvFFN]{onvolutional Feed-Forward
Network}
\newacro{cnn}[CNN]{convolutional neural network}
\newacro{dvae}[dVAE]{discrete variational autoencoder}
\newacro{coglm}[CogLM]{cross-modal general language model}
\newacro{qformer}[Q-Former]{querying transformer}
\newacro{sdgs}[SDGS]{Synthetic Data Generation System}
\newacro{sota}[SOTA]{state-of-art}
\newacro{mse}[MSE]{mean squared error}
\newacro{ssim}[SSIM]{structural similarity index measure}
\newacro{psnr}[PSNR]{peak signal-to-noise ratio}
\newacro{vssd}[VSSD]{visual state space duality}
\newacro{rag}[RAG]{retrieval-augmented generation}
\newacro{mllm}[MLLM]{multimodal large language model}
\newacro{mids}[MIDS]{multimodal image detection system}
\newacro{vae}[VAE]{variational autoencoder}
\newacro{ravid}[RAVID]{Retrieval-Augmented Visual Detection}

\definecolor{darkgreen}{RGB}{10,191,10}

\definecolor{olivegreen}{RGB}{212,232,231}

\definecolor{lightblue}{RGB}{138,170,229}
\colorlet{lightblueAlpha}{lightblue!30}

\definecolor{lightred}{RGB}{255, 194, 194}
\definecolor{iccvblue}{rgb}{0.21,0.49,0.74}
\usepackage[pagebackref,breaklinks,colorlinks,allcolors=iccvblue]{hyperref}

\def\paperID{14579} % *** Enter the Paper ID here
\def\confName{ICCV}
\def\confYear{2025}

\title{{\color{red}RAVID}: {\color{red}\underline{R}}etrieval-{\color{red}\underline{A}}ugmented {\color{red}\underline{Vi}}sual {\color{red}\underline{D}}etection: A Knowledge-Driven Approach for AI-Generated Image Identification}

\author{Mamadou Keita\\
Laboratory of IEMN, Univ. \\ Polytechnique Hauts-de-France\\
Valeneciennes, Fance\\
{\tt\small mamadou.keita@uphf.fr}
\and
Wassim Hamidouche\\
KU 6G Research Center, \\Khalifa University \\
Abu Dhabi, UAE \\
{\tt\small whamidouche@gmail.com}
\and
Hessen Bougueffa Eutamene\\
Laboratory of IEMN, Univ. \\ Polytechnique Hauts-de-France\\
Valeneciennes, Fance\\
{\tt\small Hessen.BougueffaEutamene@uphf.fr}
\and
Abdelmalik Taleb-Ahmed\\
Laboratory of IEMN, Univ. \\ Polytechnique Hauts-de-France\\
Valeneciennes, Fance\\
{\tt\small abdelmalik.taleb-ahmed@uphf.fr}
\and
Abdenour Hadid\\
Sorbonne Center for Artificial Intelligence,\\Sorbonne University, Abu Dhabi, UAE\\
{\tt\small abdenour.hadid@ieee.org}
}
\usepackage{tcolorbox} 
\begin{document}
\maketitle
\begin{abstract}
In this paper, we introduce RAVID, the first framework for AI-generated image detection that leverages visual \ac{rag}. While \ac{rag} methods have shown promise in mitigating factual inaccuracies in foundation models, they have primarily focused on text, leaving visual knowledge underexplored. Meanwhile, existing detection methods, which struggle with generalization and robustness, often rely on low-level artifacts and model-specific features, limiting their adaptability. To address this, RAVID dynamically retrieves relevant images to enhance detection. Our approach utilizes a fine-tuned CLIP image encoder, RAVID\_CLIP, enhanced with category-related prompts to improve representation learning. We further integrate a \ac{vlm} to fuse retrieved images with the query, enriching the input and improving accuracy. Given a query image, RAVID generates an embedding using RAVID\_CLIP, retrieves the most relevant images from a database, and combines these with the query image to form an enriched input for a \ac{vlm} (e.g., Qwen-VL or Openflamingo). Experiments on the UniversalFakeDetect benchmark, which covers 19 generative models, show that RAVID achieves state-of-the-art performance with an average accuracy of 93.85\%. RAVID also outperforms traditional methods in terms of robustness, maintaining high accuracy even under image degradations such as Gaussian blur and JPEG compression. Specifically, RAVID achieves an average accuracy of 80.27\% under degradation conditions, compared to 63.44\% for the state-of-the-art model C2P-CLIP, demonstrating consistent improvements in both Gaussian blur and JPEG compression scenarios. The code will be publicly available upon acceptance.
\end{abstract}  
\acresetall
% \begin{figure}[!th]
%     \centering
%     \includegraphics[width=0.8\linewidth]{figures/bar_plot2.png}
%     \caption{Comparison of accuracy (\%) between different methods (FatFormer, RINE, and RAVID) across multiple generative models. FatFormer adapts CLIP’s vision-language space to detect forgery traces in image and frequency domains. RINE maps CLIP features to a forgery-aware space for improved detection. In contrast, RAVID leverages \acs{rag} for enhanced generalization.}
%     \label{fig:comparison}
% \end{figure}

\begin{figure}[!ht]
    \centering
    \includegraphics[width=\linewidth]{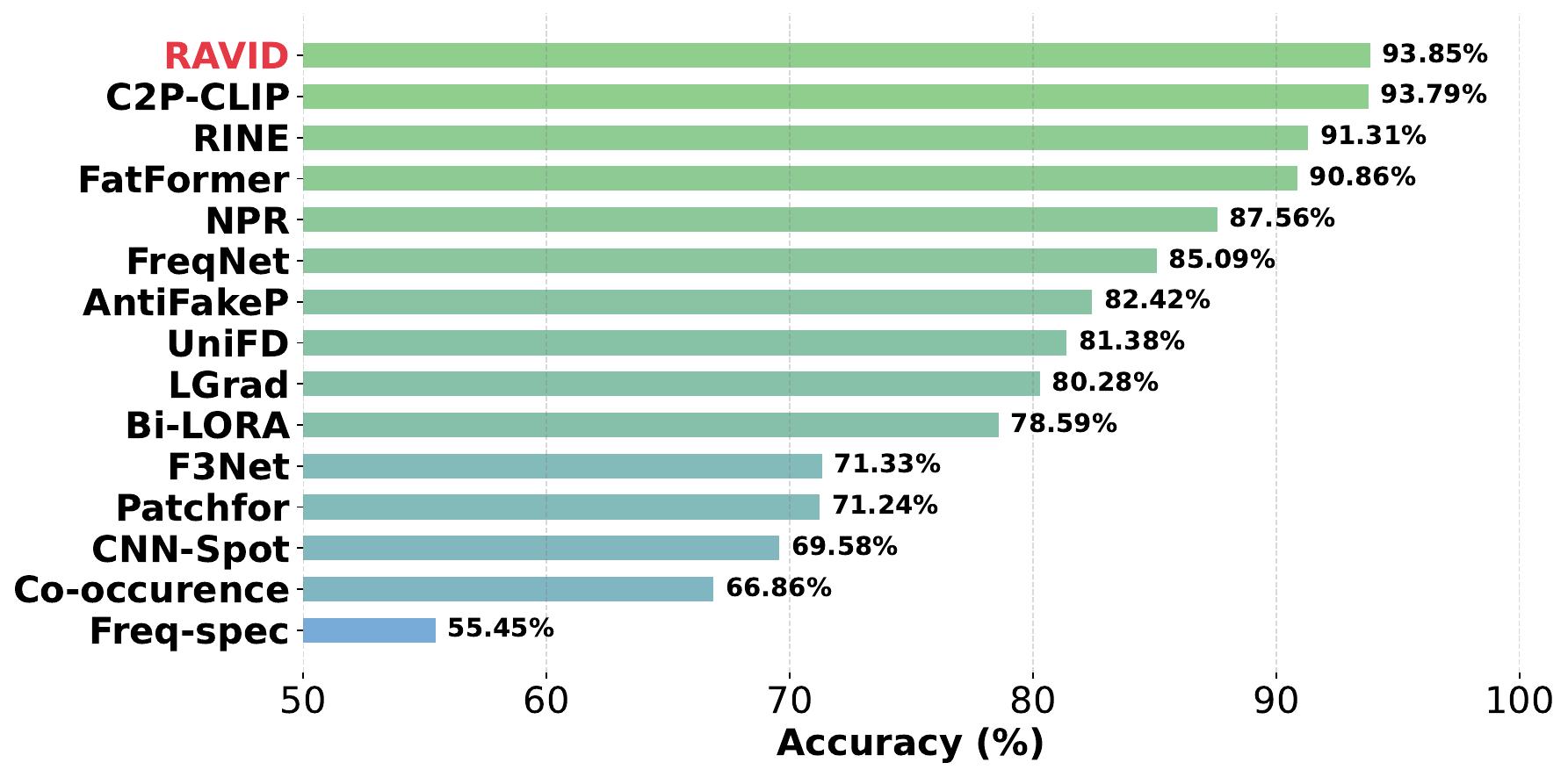}
    \caption{Performance comparison of detectors on the UniversalFakeDetect dataset, including frequency-based (Freq-spec, Co-occurrence), convolutional (CNN-Spot, F3Net), transformer-based (FatFormer, RINE, C2P-CLIP),  hybrid (Bi-LORA, LGrad), multimodal (AntifakePrompt, Bi-LORA) architectures and RAG (RAVID (ours)).}
    \label{fig:acc}
\end{figure}

%\begin{figure}[!ht]
    %\centering
    %\includegraphics[width=\linewidth]{figures/High_level.png}
    %\caption{RAVID.}
    %\label{fig:ravid}
%\end{figure}

\vspace{-2mm}
\section{Introduction}
\label{sec:intro}
The rapid advancement of generative models, particularly in image synthesis, has introduced significant challenges in distinguishing AI-generated content from real data. 
For instance, \acp{gan}~\cite{karras2018progressive,karras2019style,choi2018stargan} and diffusion-based models~\cite{nichol2022glide,dhariwal2021diffusion,rombach2022high,ramesh2021zero} have become increasingly proficient at producing photorealistic images that are nearly indistinguishable from genuine ones. However, the progress in detection methods has not kept pace with these advancements, creating an urgent need for robust and reliable detection systems. As generative models continue to evolve, so do the complexities associated with their detection, leading to a constant race between improving generative techniques and enhancing detection capabilities.

Traditional AI-generated image detection approaches primarily rely on identifying low-level artifacts or model-specific fingerprints~\cite{sinitsa2024deep} present in synthetic images. These artifacts include pixel inconsistencies, noise patterns, and subtle distortions that reveal traces of the underlying generation process. While these methods have demonstrated effectiveness in controlled settings, they often fail in real-world scenarios. As generative models improve, they become more adept at minimizing artifacts and replicating the statistical properties of real images, making detection increasingly difficult. Furthermore, many existing detection methods suffer from a fundamental limitation: over-reliance on model-specific features and low-level artifacts. Since these methods are often tailored to exploit weaknesses in particular architectures, they struggle to generalize across different generative models. Consequently, there is a growing need for more adaptive detection approaches that leverage additional sources of information to enhance performance, robustness, and reliability.

One promising direction is \ac{rag}~\cite{lewis2020retrieval}, a paradigm initially developed to improve factual accuracy in large language models by retrieving and incorporating external knowledge relevant to a given query. While extensively explored in textual tasks, its potential for visual tasks, particularly AI-generated image detection, remains largely underexplored. Existing detection methods mainly rely on hand-crafted features, model fingerprints, and deep learning-based classifiers trained on limited datasets~\cite{wang2023dire,cozzolino2024raising}. These approaches face significant challenges in generalization, struggling to detect images generated by unseen models, and robustness, as even minor perturbations, such as noise injection, compression artifacts, or adversarial attacks, can significantly degrade their performance. Overcoming these limitations requires a more adaptive, retrieval-based framework that dynamically integrates external visual knowledge to improve decision-making.

To address this gap, we propose RAVID, a novel retrieval-augmented framework for AI-generated image detection. Unlike traditional methods that rely solely on model-dependent features, RAVID retrieves visually similar images relevant to the input query and integrates them into the detection process, thereby enhancing accuracy and robustness. At its core, RAVID leverages a CLIP-based image encoder, fine-tuned through category-level prompt integration to improve its ability to capture semantic and structural patterns crucial for distinguishing AI-generated images from real ones. Additionally, we incorporate \acp{vlm}, such as Openflamingo~\cite{awadalla2023openflamingo}, to effectively fuse retrieved images with the query input, enabling richer contextual understanding. By combining retrieval-based augmentation with advanced vision-language decision-making ability, our approach significantly improves adaptability and effectiveness, making it well-suited for real-world applications where reliable AI-generated content detection is critical. To evaluate our approach, we conduct extensive experiments on UniversalFakeDetect, a large-scale benchmark comprising AI-generated images from 19 generative models. This diverse dataset allows for rigorous assessment of our framework’s generalization and robustness across in- and out-of-domain scenarios. Experimental results demonstrate that RAVID consistently outperforms existing detection techniques, achieving an average accuracy of 93.85\%  and exhibiting better generalization across multiple challenging settings. As depicted in Figure~\ref{fig:acc}, RAVID achieves the highest accuracy, outperforming all compared state-of-the-art detectors. In addition, RAVID also demonstrates strong resilience to image degradations, achieving 80.27\% average accuracy under conditions such as Gaussian blur and JPEG compression significantly outperforming the state-of-the-art C2P-CLIP model, which achieves 63.44\% under the same settings. Unlike traditional methods, it maintains high accuracy by leveraging retrieval-augmented generation to compensate for lost visual features, ensuring robust performance even under real-world distortions. This highlights the effectiveness of retrieval-augmented techniques in enhancing generalization and robustness across different image generators. Our key contributions are summarized as follows:
\begin{itemize}
\item A novel retrieval-augmented framework for AI-generated image detection, dynamically retrieving and integrating external visual knowledge to enhance decision-making.
\item Fine-tuning of a CLIP-based image encoder with category-level prompt integration, improving representation learning for retrieval tasks.
\item Integration of \acp{vlm} to combine retrieved images with queries, enhancing contextual understanding and detection robustness.
\item Comprehensive evaluation on UniversalFakeDetect, demonstrating significant improvements in generalization and robustness over existing methods.
\end{itemize}

The rest of this paper is structured as follows. Section~\ref{sec:relatedWork} reviews related work on AI-generated image detection, AI image generation, and visual \ac{rag}. Section~\ref{sec:method} introduces the proposed visual \ac{rag}-based approach for AI-generated image identification. Section~\ref{sec:experiments} evaluates and analyzes the performance of the proposed detection framework. Finally, Section~\ref{sec:conclusion} summarizes the findings and concludes the paper.

\begin{figure*}[!th]
    \centering
    \includegraphics[width=0.9\linewidth]{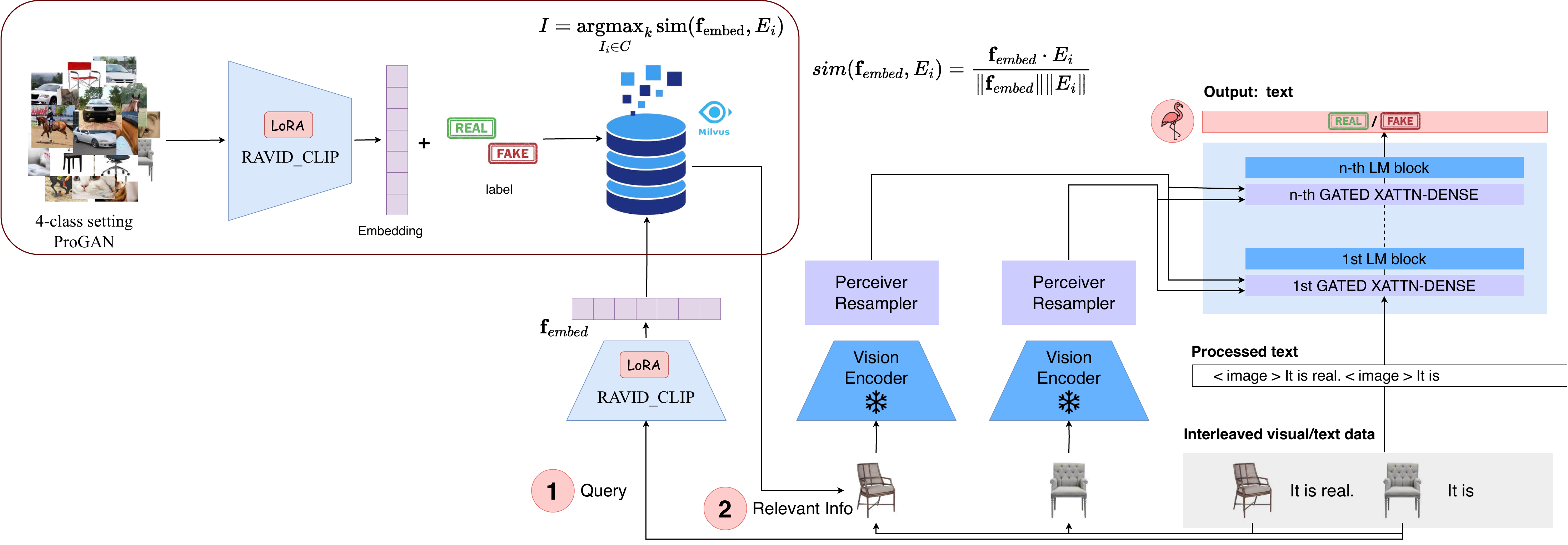}
    \caption{Our RAVID integrates RAVID\_CLIP for embedding-based image retrieval and Openflamingo for decision-making: (1) 4-class ProGAN training set images are encoded into vector embeddings using RAVID\_CLIP and stored in a Milvus vector database; (2) At testing time, the query image embedding is matched against stored embeddings to retrieve the most relevant images; (3) The retrieved images and labels serve as contextual information, combined with the query image, and processed by Openflamingo.}
    \label{fig:method}
\end{figure*}

\section{Related Works}
\label{sec:relatedWork}
\subsection{Image Generation}
Recent advances in deep generative models have significantly enhanced the capabilities of synthetic image generation. Early breakthroughs included the introduction of \acp{gan} by Goodfellow {\it et al.}~\cite{goodfellow2014generative}, enabling the creation of realistic images without relying on input data. Subsequent research refined \acp{gan} to improve image quality and diversity, and to introduce conditional generation, leading to models like ProGAN ~\cite{karras2018progressive}, StyleGAN~\cite{karras2019style}, BigGAN~\cite{brocklarge}, CycleGAN~\cite{zhu2017unpaired}, StarGAN~\cite{choi2018stargan}, and GauGAN~\cite{park2019gaugan}. More recently, diffusion models have gained prominence, particularly for text-to-image synthesis. Models like Glide~\cite{nichol2022glide}, \ac{ldm}~\cite{rombach2022high}, \ac{adm}~\cite{dhariwal2021diffusion}, DALL-E 3~\cite{dalle3_openai}, Midjourney v5~\cite{midjourney_v5}, Firefly~\cite{adobe_firefly}, Imagen~\cite{saharia2022photorealistic}, SDXL~\cite{podell2023sdxl}, and SGXL~\cite{sauer2022stylegan} have demonstrated impressive quality and versatility across diverse categories and scenes, often surpassing \acp{gan}-based models in flexibility and image fidelity. 

\subsection{Visual Retrieval-Augmented Generation}
\Ac{rag} enhances foundation models by integrating external knowledge into the generation process. It comprises two key components: retrieval and generation. Given a query $q$, the retrieval module selects relevant documents $K$ from an external knowledge source (e.g., Wikipedia) based on similarity measures. These retrieved elements serve as additional input to the generation module, which then produces the final response $y$ using a causal model such as a \ac{llm}. Despite its effectiveness in text-based tasks, traditional RAG models primarily focus on textual data, leaving visual knowledge largely unexplored. Recently, a few studies have begun exploring visual \ac{rag}~\cite{yu2024visrag,riedler2024beyond}. For instance, Riedler {\it et al.}~\cite{riedler2024beyond} investigated multimodal \ac{rag} in industrial applications, assessing image-text integration through multimodal embeddings and textual summaries generated by GPT-4V and LLaVA~\cite{liu2023visual}. Their findings indicate that while textual summaries improve performance, image retrieval remains a significant challenge. To address these limitations, {\it Yu et al.}~\cite{yu2024visrag} introduced VisRAG, a \ac{vlm}-based \ac{rag} framework that encodes documents as images for retrieval. By preserving visual and layout information, VisRAG enhances generation quality and outperforms text-based \ac{rag} by 20 to 40\%. Similarly, Ren {\it et al.}~\cite{ren2025videorag} proposed VideoRAG, a framework designed for long-context video understanding. Their approach integrates graph-based textual grounding and multimodal context encoding, efficiently preserving both semantic relationships and visual features. While existing video-integrated \ac{rag} methods often predefine query-related videos or convert them into text, thereby losing multimodal richness, Jeong {\it et al.}~\cite{jeong2025videorag} proposed an improved VideoRAG framework. Their approach dynamically retrieves relevant videos and leverages \acp{vlm} to integrate visual and textual data, enabling enhanced multimodal response generation.

\subsection{Detection Methods}
Detecting AI-generated images has become increasingly critical with the proliferation of synthetic content. As image generation methods evolve, researchers have developed various techniques to enhance detection accuracy and generalization. These approaches can be broadly categorized into three domains: spatial, frequency, and multimodal features. In the spatial domain, Wang {\it et al.}~\cite{wang2023dire} demonstrated that genuine images exhibit higher \ac{ddim} inversion errors than AI-generated images. In the frequency domain, Jeong {\it et al.}~\cite{jeong2022bihpf, jeong2022fingerprintnet, jeong2022frepgan} explored techniques leveraging frequency-based artifacts for improved detection. Meanwhile, multimodal methods have shown promise in enhancing robustness and generalization across diverse datasets. For instance, Chang {\it et al.}~\cite{chang2023antifakeprompt} introduced a zero-shot deepfake detection method using \acp{vlm} like InstructBLIP. By framing detection as a \ac{vqa} task, they improved accuracy on unseen data through prompt tuning. Similarly, Keita {\it et al.}~\cite{keita2025bi} proposed Bi-LORA, which reframes binary classification as an image captioning problem, leading to high-precision synthetic image detection. Other studies have focused on integrating category-specific information to enhance feature extraction. Tan {\it et al.}~\cite{tan2024c2p} developed C2P-CLIP, which embeds category-specific concepts into CLIP’s image encoder, achieving state-of-the-art generalization in deepfake detection. Huang {\it et al.}~\cite{huang2024ffaa} introduced OW-FFA-VQA, leveraging a \ac{vqa} framework and the FFA-VQA dataset for explainable face forgery analysis using a fine-tuned \ac{mllm} and \ac{mids}. Furthermore, Iliopoulou {\it et al.}~\cite{iliopoulou2024synthetic} explored compression-based detection methods using a \ac{vae}, distinguishing real and synthetic faces based on reconstruction quality. Koutlis {\it et al.}~\cite{koutlis2024leveraging} further improved synthetic image detection by mapping intermediate CLIP Transformer block features to a forgery-aware vector space, achieving a +10.6\% performance improvement over previous state-of-the-art methods with minimal training. 

Despite the substantial progress in AI-generated image detection, none of the existing methods have explored the integration of \ac{rag} within \acp{vlm} for this task. Given \ac{rag}’s proven success in various domains, its potential for enhancing AI-generated image detection remains entirely unexplored. Addressing this critical gap, we introduce RAVID, the first \ac{rag}-driven \ac{vlm} framework for AI-generated image detection, leveraging retrieved visual knowledge to improve detection accuracy, robustness, and generalization across diverse generative models.

\begin{figure*}
    \centering
    \begin{subfigure}[b]{0.24\linewidth}
        \includegraphics[width=\linewidth]{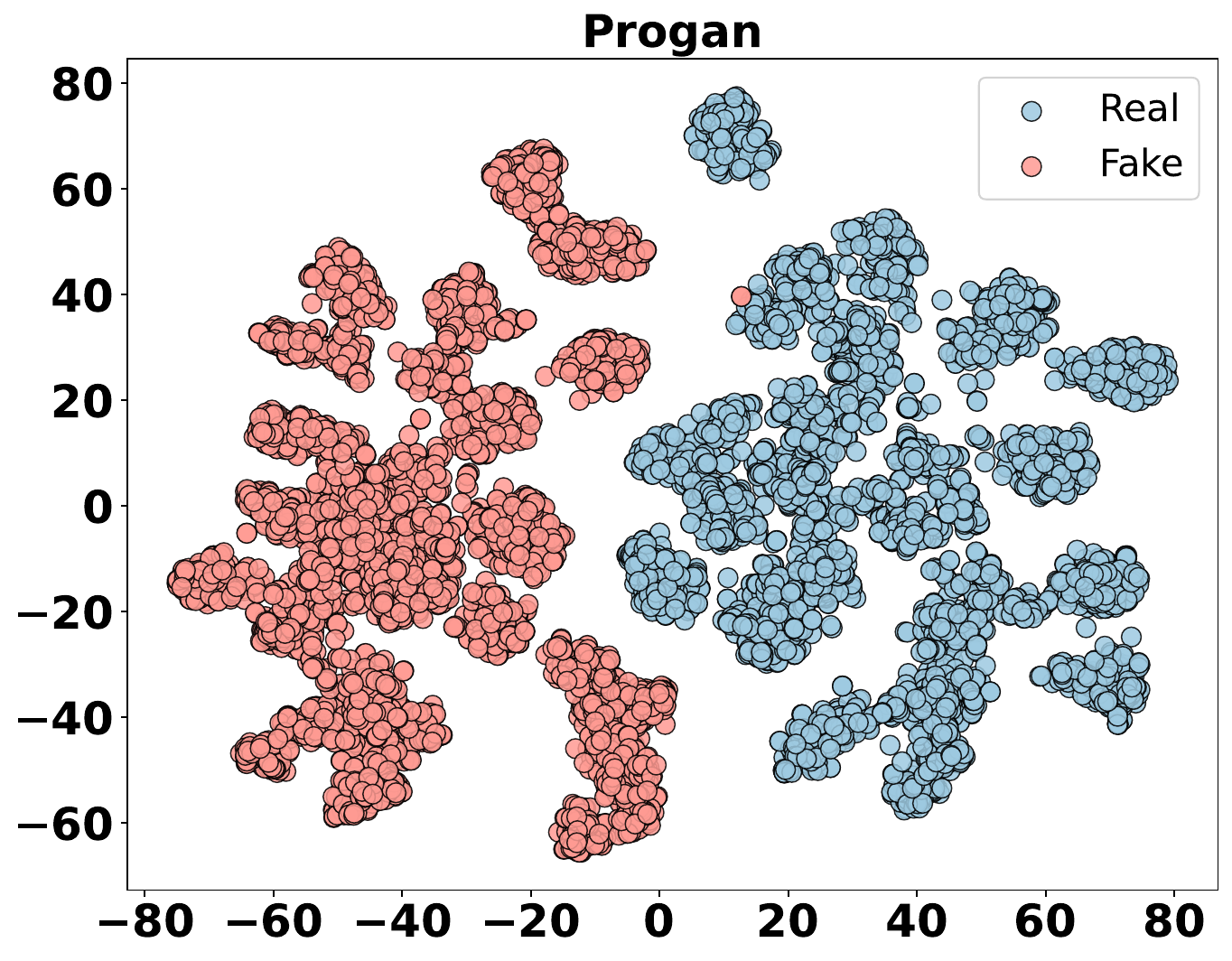}
        %\caption*{progan \hspace{8mm} }
    \end{subfigure}
%    \hspace{.1in}
   \vspace{.001in}
    \begin{subfigure}[b]{0.24\linewidth}
        \includegraphics[width=\linewidth]{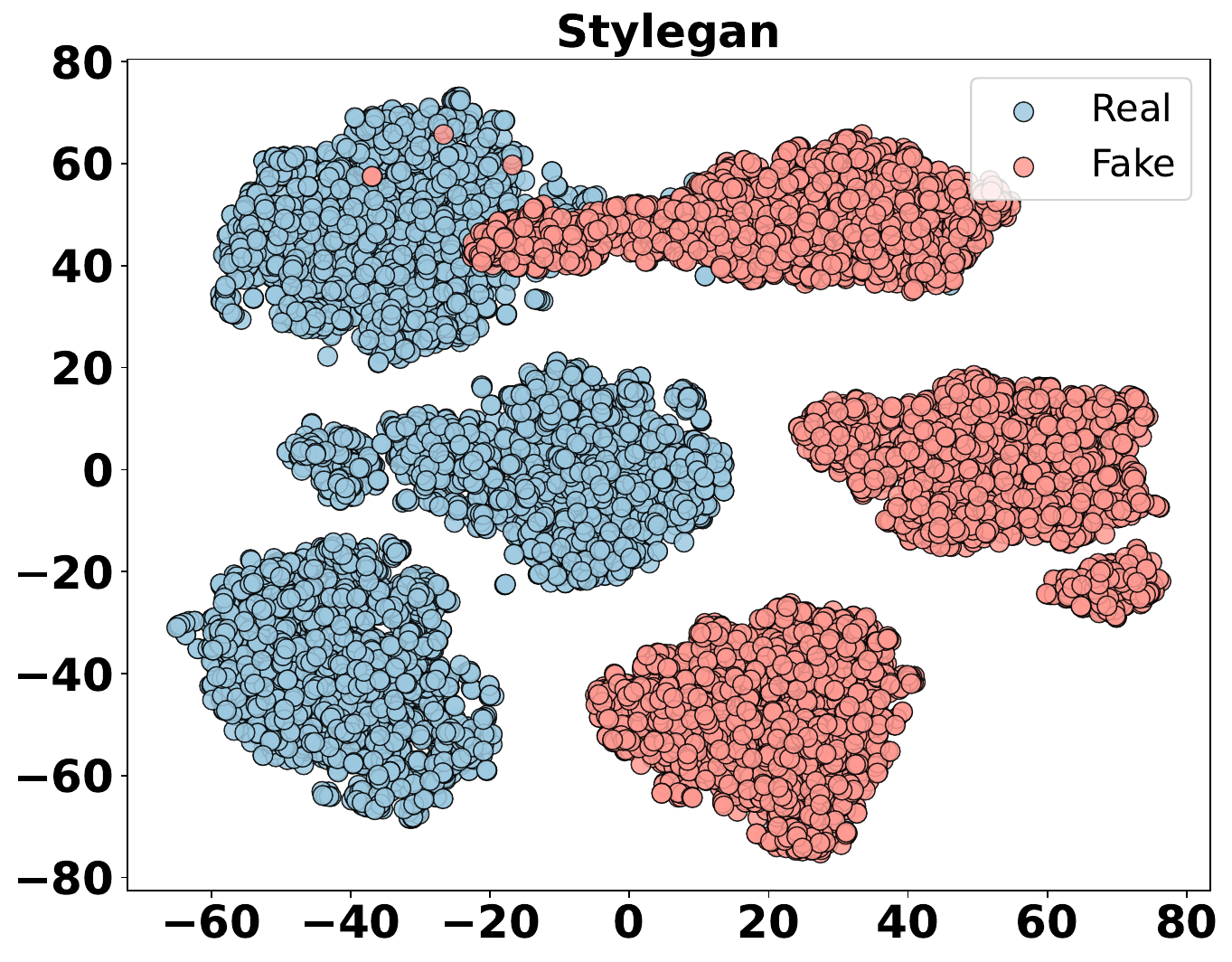}
        %\caption*{stylegan \hspace{15mm}}
    \end{subfigure}
   % \hspace{.1in}
   \vspace{.001in}
    \begin{subfigure}[b]{0.24\linewidth}
        \includegraphics[width=\linewidth]{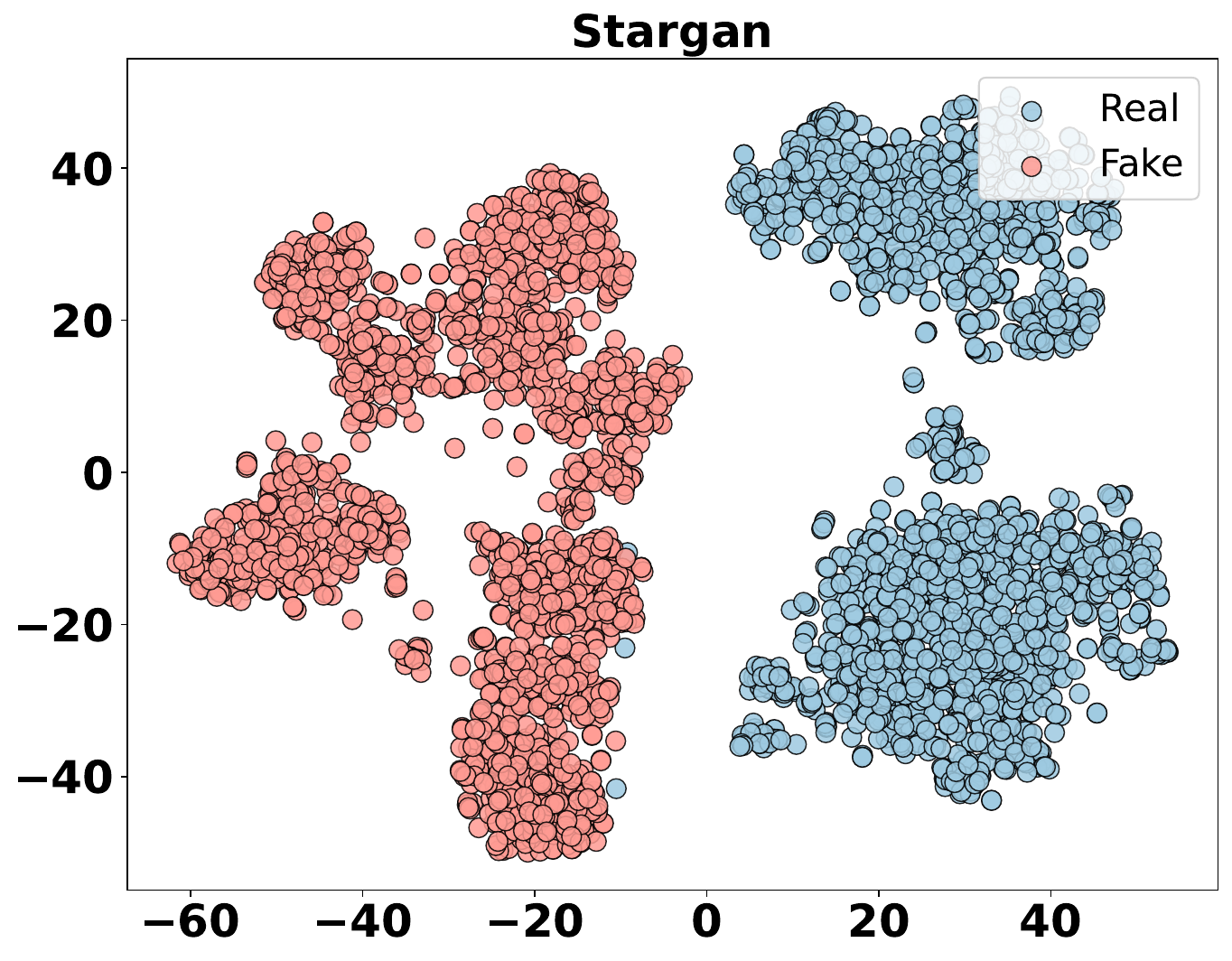}
        %\caption*{stargan \hspace{5mm}}
    \end{subfigure}
   % \hspace{.1in}
   \vspace{.001in}
    \begin{subfigure}[b]{0.24\linewidth}
        \includegraphics[width=\linewidth]{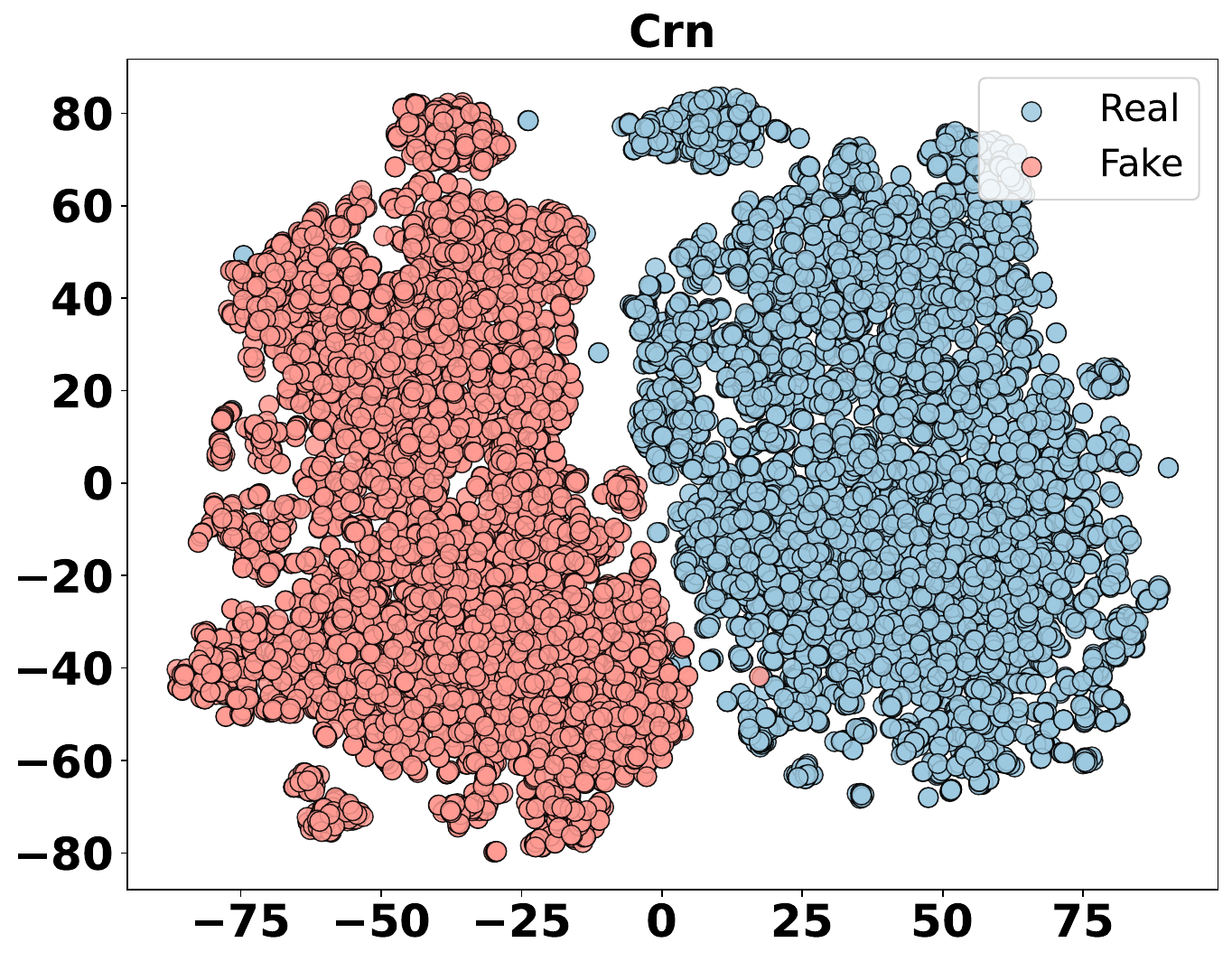}
        %\caption*{crn\hspace{5mm}}
    \end{subfigure}
    \begin{subfigure}[b]{0.24\linewidth}
\includegraphics[width=\linewidth]{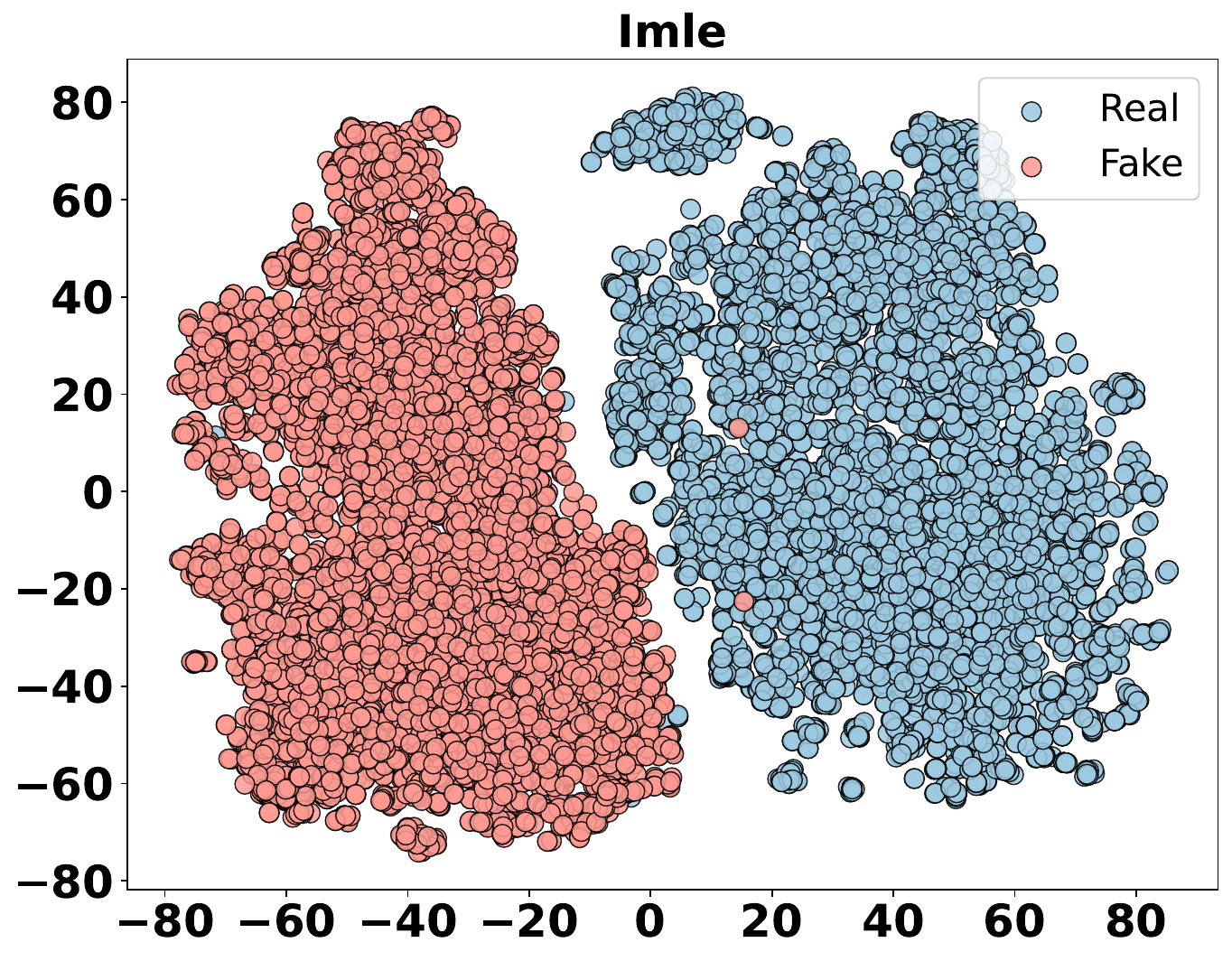}
        %\caption*{imle}
    \end{subfigure}
  %  \hspace{.1in}
   \vspace{.001in}
    \begin{subfigure}[b]{0.24\linewidth}
        \includegraphics[width=\linewidth]{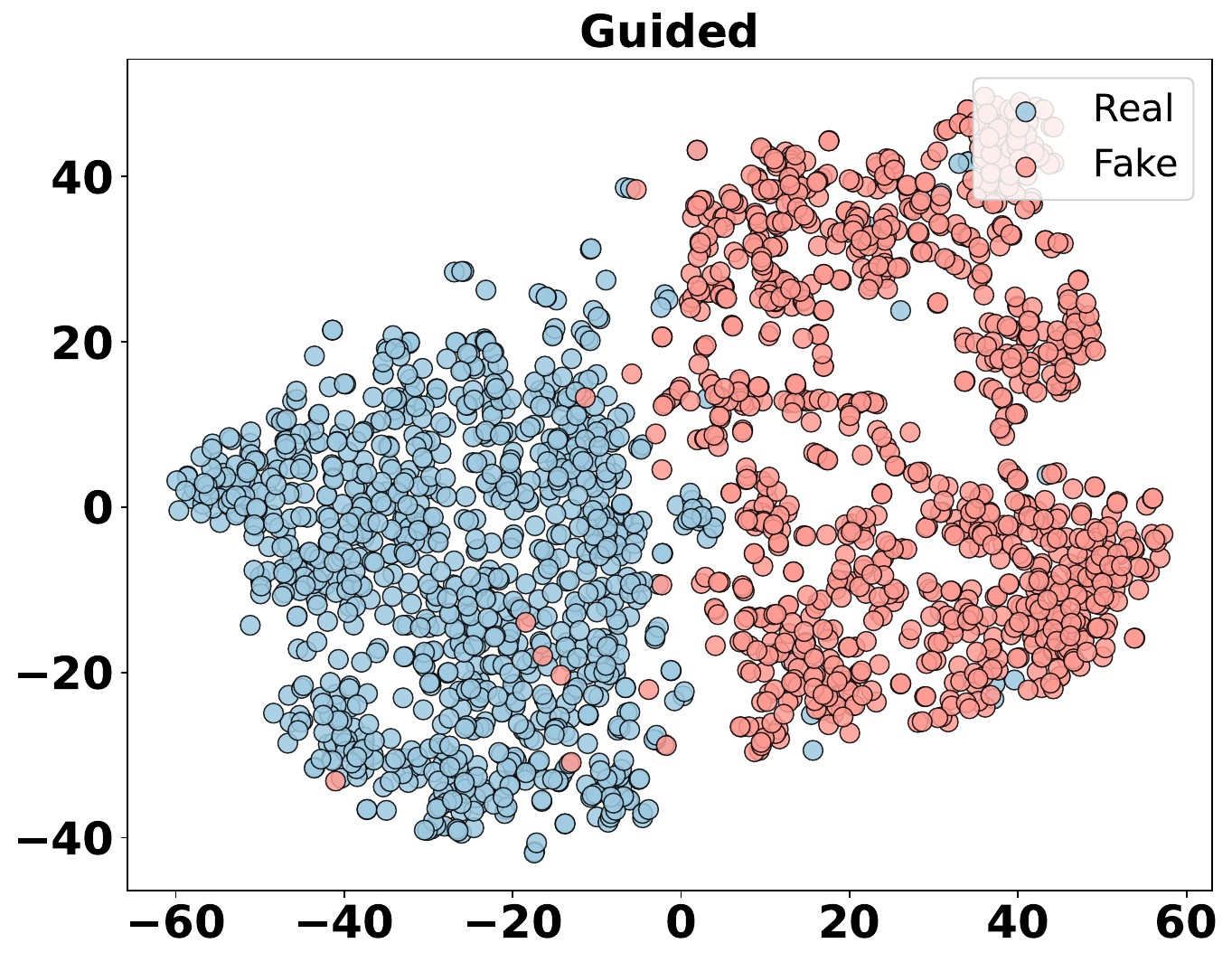}
        %\caption*{guided}
    \end{subfigure}
    % \hspace{.1in}
    \vspace{.001in}
    \begin{subfigure}[b]{0.24\linewidth}     \includegraphics[width=\linewidth]{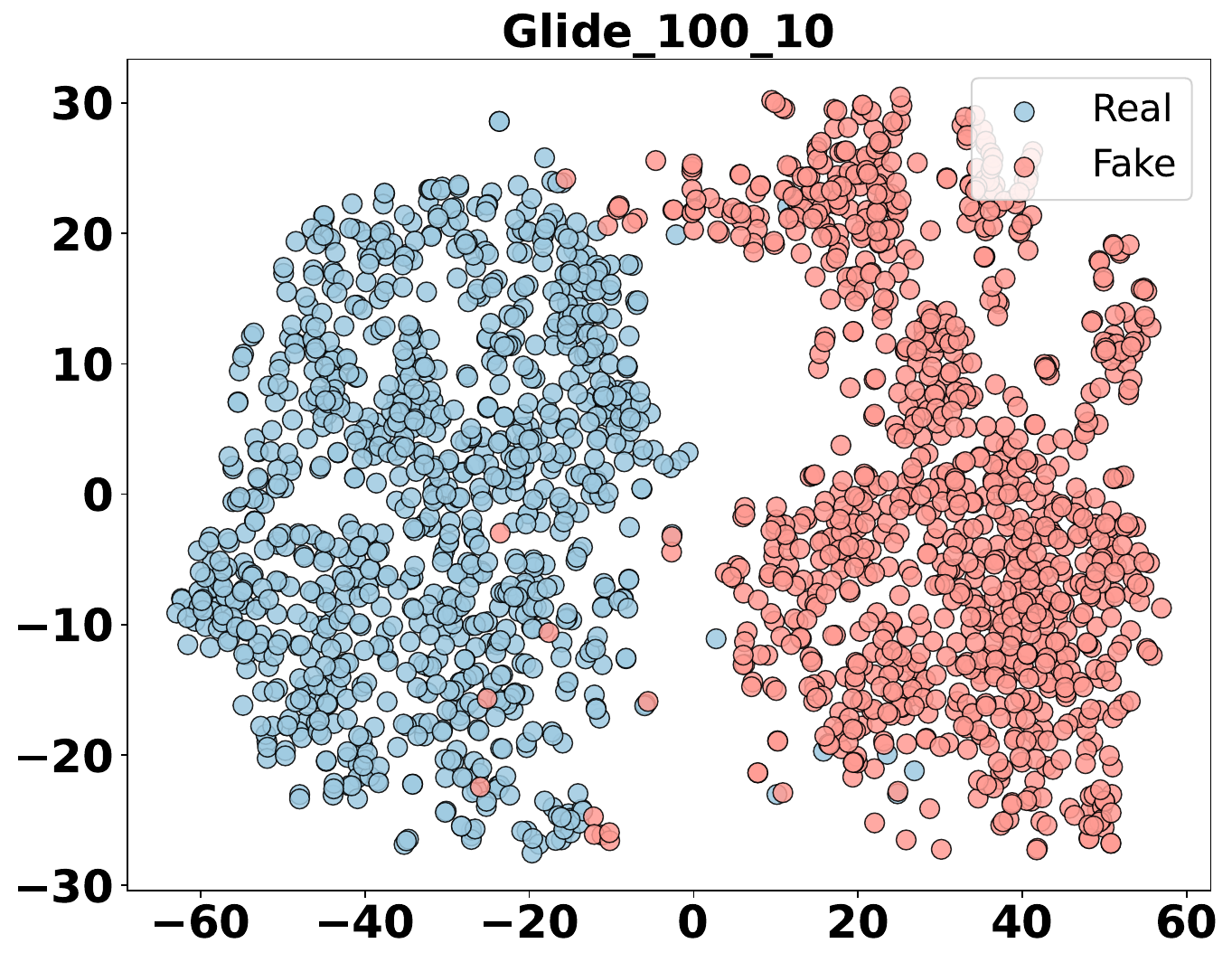}
        %\caption*{glide\_100\_10}
    \end{subfigure}
%    \hspace{.1in}
   \vspace{.001in}
     \begin{subfigure}[b]{0.24\linewidth}     \includegraphics[width=\linewidth]{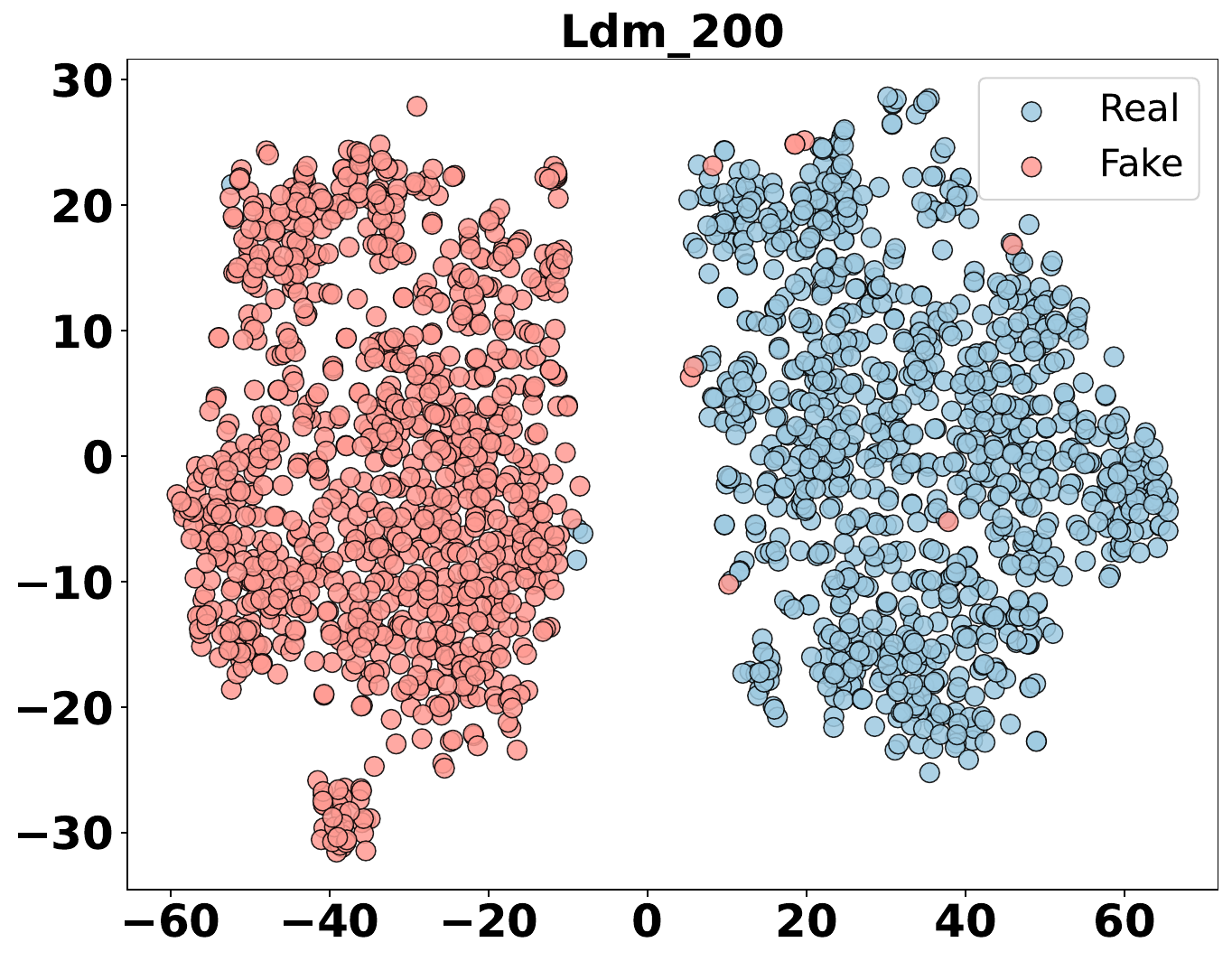}
         %\caption*{ldm\_200}
     \end{subfigure}  
    \caption{t-SNE Visualization of RAVID\_CLIP's Feature Distributions for Different Generative Models. The scatter plots illustrate the t-SNE embeddings of features extracted from real (green) and generated (red) images across various generative models.}
    \label{fig:Visual} 
\end{figure*}

\section{Proposed Method}
\label{sec:method}
In this section, we present RAVID, which performs retrieval of query-relevant images over vector image corpus as the external knowledge source and determines query-image class (label) grounded in them. 

% The detailed procedure of RAVID is summarized in Algorithm~\ref{alg:RAVID_Algo}.

\subsection{Motivation}
As generative models continue to improve, distinguishing AI-generated images from real ones has become increasingly challenging. Traditional detection methods often rely on low-level artifacts or model-specific fingerprints, making them prone to overfitting and inconsistent performance across different synthesis techniques. These methods also struggle with generalization and lack robustness, as even small changes can significantly impact detection accuracy. Recent research often focuses on using \acp{vlm} for AI-generated image detection. However, most of these approaches require extensive fine-tuning of \acp{vlm}, which is computationally expensive, limits scalability, and can lead to overfitting on specific datasets.

To address these challenges, we propose RAVID, a novel approach that combines \ac{rag} with AI-generated image detection. By integrating external image retrieval, RAVID dynamically enhances detection by providing relevant context, overcoming the limitations of existing methods without the need for \ac{vlm} fine-tuning. Our approach improves the robustness of AI-generated image detection, ensuring higher accuracy and better generalization in diverse scenarios by incorporating large-scale visual knowledge through a vector image database. With RAVID, we present a novel approach that combines powerful image embeddings with visual-language models, without the need for fine-tuning. This method sets a new standard for detecting AI-generated images, offering improved performance across a wide variety of generative models.

%While generative models continue to advance, distinguishing AI-generated images from real ones has become increasingly difficult. Traditional detection methods typically rely on low-level artifacts or model fingerprints, making them vulnerable to overfitting and inconsistent performance across different synthesis techniques. Additionally, these methods often struggle with generalization, as even minor perturbations can significantly degrade detection accuracy. A common approach in recent research is to leverage \acp{vlm} for AI-generated image detection. However, most existing works require extensive fine-tuning of the \acp{vlm}, which is computationally expensive, limits scalability, and risks overfitting specific datasets. To address these challenges, we propose RAVID, which integrates Retrieval-Augmented Generation (RAG) with AI-generated image detection. Our approach leverages the power of external image retrieval to dynamically enhance the analysis by providing relevant context and overcoming the limitations of existing methods without requiring VLM fine-tuning. We aim to enhance the robustness of AI-generated image detection, ensuring higher accuracy and better generalization in diverse scenarios by incorporating large-scale visual knowledge through a vector image database. Through RAVID, we introduce a novel method that combines advanced image embeddings and large vision-language models without fine-tuning to achieve superior performance, setting a new standard for detecting AI-generated images across various models.

\subsection{Concept-Aware Image Embeddings}

Recent work by Tan \textit{et al.}~\cite{tan2024c2p} demonstrates that CLIP features' ability to detect AI-generated images through linear classification is largely due to their capacity to capture \textit{conceptual similarities}. Building on this insight, they propose an approach that incorporates enhanced captions and contrastive learning to embed categorical concepts into the image encoder, thereby improving the distinction between real and generated images. Inspired by this work, we adopt a similar strategy in our framework to generate high-quality embeddings for a vector database, which is crucial to the performance of the RAVID method.

\begin{enumerate}
    \item \textbf{Caption Generation and Enhancement}:\\

Let \( \mathcal{D} \) represent the training dataset containing both real and synthetic images, defined as \( \mathcal{D} = \{(x_j, y_j)\}_{j=1}^{N} \), where \( y_j \in \{0, 1\} \) indicates whether the image \( x_j \) is real (\( y = 0 \)) or fake (\( y = 1 \)). For each image in the dataset, we generate captions using the ClipCap model~\cite{mokady2021clipcap}, resulting in a set of captions \( \mathcal{C} = \{(c_j, y_j)\}_{j=1}^{N} \).

To enhance these captions, we append category-specific prompts \( \mathcal{P} = \{P_{\text{real}}, P_{\text{fake}}\} \) to the original captions, as proposed by~\cite{tan2024c2p}. Specifically, the enhanced captions \( \tilde{\mathcal{C}} = \{\tilde{c}_j\}_{j=1}^{N} \) are defined as:

\begin{equation}
\tilde{c}_j =
\begin{cases}
(P_{\text{real}}, c_j), & \text{if } y_j = 0 \\
(P_{\text{fake}}, c_j), & \text{if } y_j = 1
\end{cases}
\end{equation}

In this formulation, \( P_{\text{real}} \) and \( P_{\text{fake}} \) represent category-specific prompts (e.g., \( P_{\text{real}} = \text{Camera} \), \( P_{\text{fake}} = \text{Deepfake} \)) that provide additional context to differentiate real images from synthetic ones.\\

\item \textbf{Concept Injection via Contrastive Learning}:\\

To integrate classification concepts into the image encoder, we employ a contrastive learning framework. In this approach, the text encoder remains frozen, while Low-Rank Adaptation (LoRA) layers are applied to the image encoder to facilitate learning. Given an image \( x_j \) and its corresponding enhanced caption \( \tilde{c}_j \), their feature representations are computed as follows:

\begin{equation}
    t_j = \text{encoder}_{\text{text}}(\tilde{c}_j), \quad \mathbf{e}_j = \text{encoder}_{\text{img}}(x_j),
\end{equation}

where \( t_j \) and \( \mathbf{e}_j \) denote the text and image embeddings, respectively.

To ensure that the image encoder aligns visual features with their corresponding textual descriptions, we optimize a contrastive loss function \( L_{\text{contrastive}} \), defined as:

\begin{equation}
    L_{\text{contrastive}} = \frac{1}{2} \left( L_{\mathbf{e} \to t} + L_{t \to \mathbf{e}} \right),
\end{equation}

where \( L_{\mathbf{e} \to t} \) enforces alignment between image embeddings and their respective text embeddings, while \( L_{t \to \mathbf{e}} \) ensures reverse alignment. These losses are formulated as:

\begin{equation}
    L_{\mathbf{e} \to t} = -\frac{1}{N} \sum_{i=1}^{N} \log \frac{\exp(\mathbf{e}^T_i \cdot t_i)}{\sum_{j=1}^{N} \exp(\mathbf{e}^T_i \cdot t_j)},
\end{equation}

\begin{equation}
    L_{t \to \mathbf{e}} = -\frac{1}{N} \sum_{i=1}^{N} \log \frac{\exp(t^T_i \cdot \mathbf{e}_i)}{\sum_{j=1}^{N} \exp(t^T_i \cdot \mathbf{e}_j)}.
\end{equation}

Here, \( \mathbf{e}_i^T t_j \) represents the dot product between the image feature \( \mathbf{e}_i \) and the text feature \( t_j \), capturing their similarity. The denominator normalizes the similarity scores across all samples in the batch, ensuring a well-structured representation space.

By optimizing \( L_{\text{contrastive}} \), the image encoder learns to map visual features into a space that aligns with their textual descriptions. This process effectively injects classification concepts within the image encoder, enhancing its ability to distinguish between real and AI-generated images.
\end{enumerate}

\noindent The CLIP concept-injection-enhanced model is then used to generate embeddings for both real and fake images. These embeddings are stored in a vector database (e.g., Milvus~\cite{milvus2025}), which serves as an external knowledge source for the RAVID framework.

\subsection{Image Retrieval}

Retrieval involves computing the similarity between the query image \( q_{\text{img}} \) and each knowledge element (image embeddings) to determine relevance. To achieve this, we first embed the query image \( q_{\text{img}} \) using the RAVID\_CLIP image encoder to obtain its embedding \( \mathbf{f}_{\text{embed}} \). Relevance is then computed based on representation-level similarity, such as cosine similarity, to measure the alignment between the query embedding and stored embeddings in the external corpus \( C \). The retrieval process is formulated as:

\begin{equation}
    I = \underset{I_i \in C}{\text{argmax}_k} \, \text{sim}(\mathbf{f}_{\text{embed}}, E_i),
\end{equation}

where \( \mathbf{f}_{\text{embed}} \) is the embedding of the query image \( q_{\text{img}} \) computed by the RAVID\_CLIP image encoder, \( E_i \) represents the stored embedding of an image \( I_i \) in the external corpus \( C \), and \( \text{sim}(\mathbf{f}_{\text{embed}}, E_i) \) denotes the cosine similarity between the query embedding and each corpus embedding, computed as:

\begin{equation}
    \text{sim}(\mathbf{f}_{\text{embed}}, E_i) = \frac{\mathbf{f}_{\text{embed}} \cdot E_i}{\|\mathbf{f}_{\text{embed}}\| \|E_i\|}
\end{equation}

where \( \underset{I_i \in C}{\text{argmax}_k} \) selects the top-\( k \) images with the highest similarity scores.

By retrieving the top-\( k \) most relevant images, this approach ensures that the subsequent answer generation step benefits from rich contextual information, improving the robustness of AI-generated image detection.

%The first step in operationalizing RAG over the image embedding corpus is image retrieval, which aims to identify query-relevant images \( I = \{ I_1, I_2, \dots, I_k \} \) from an external corpus \( C \) containing a large number of image embedding, using cosine similarity. This process is formalized as:

%\[
%retrieved_images = \text{Retriever}%(f_{embed}, nb, C)
%\]

%Here, retrieval involves computing the similarity between the query image \( q_{img} \) and each knowledge element (image embeddings) to determine relevance. To achieve this, we first embed the query image \( q_{img} \) using RAVID\_CLIP image encoder to obtain its embedding \( f_{embed} \). Next, relevance is computed based on representation-level similarity—such as cosine similarity—and the top-k most similar images are retrieved for the subsequent answer generation step.

%information such as \( img_{path} \) and \( label \)

\subsection{Image-Augmented Response Generation}
After retrieving the most relevant images, the next step is to integrate them into the response generation process to enhance the quality and contextual accuracy of the generated output. To achieve this, we first construct a multimodal context by pairing each retrieved image with its corresponding label. These multimodal pairs are then concatenated to form a comprehensive context representation. Finally, the query image is incorporated into this structured input, which serves as the input to a \ac{vlm}, such as Openflamingo. Formally, this process is represented as:\\

%\subsection{Image-Augmented Response Generation}

%Once the relevant images have been retrieved, the next step is to incorporate the retrieved images into the response generation process, in order to formulate the response based on them. To make this operational, we first prepare a context with each retrieved image and corresponding label, then concatenate these multimodal pairs for all retrieved images, and finally add the query image, to construct the LVLM (eg. Qwen-VL) input, denoted as follows: 
\begin{figure}[t]
    \centering
    \resizebox{0.8\columnwidth}{!}{
    \begin{tcolorbox}[colframe=blue!50!black, colback=blue!5, sharp corners, boxrule=0.8pt, width=0.95\linewidth, title= LLM Prompt for AI-Generated Image Detection, fonttitle=\small]
    \small
    \textbf{Role:} You are an AI-generated image Detection System that leverages visual retrieval-augmented generation to enhance accuracy and robustness through multimodal context fusion. \\[3pt]
    \textbf{Task:} Identify whether a given image is AI-generated or real by retrieving relevant visual references and analyzing them alongside the query image using a vision-language model. \\[3pt]
    \textbf{Objective:} Achieve SOTA performance in AI-generated image detection with strong generalization across diverse generative models and high robustness under image degradations. \\[3pt]
    \textbf{Constraints:}
    \begin{itemize}
        \item Answer the question with a single word.
        % \item Minimize RAM usage by reducing intermediate activation size.
        % \item Prioritize stride=2 in early blocks for downsampling.
        % \item Use smaller expansion\_factor and output\_channels in early blocks.
        % \item Limit SE blocks and their ratios to reduce activation memory.
        % \item Ensure total MACs $\leq$ 350M.
        % \item Image size: 160$\times$160.
    \end{itemize}
    % \textbf{Search Space:} Use only values from the hierarchical search space: json\_search\_space.
    \textbf{Search Space:} Use only knowledge from the additional context in decision-making.
    \end{tcolorbox}
    }
    \caption{Example of initial prompt used to guide LLM architecture generation.}
    \label{fig:llm_prompt} \vspace{-4mm}
\end{figure}

\begin{lstlisting}
[
    {'text': 'Is this photo real? Please provide your answer. You should ONLY output "real" or "fake".'},
    {'image': 'path to img_1'},
    {'text': 'User: It is \nAssistant: img_1_label'},
    {'image': 'path to img_2'},
    {'text': 'User: It is \nAssistant: img_2_label'},
    ...
    {'image': 'path to img_n'},
    {'text': 'User: It is \nAssistant: img_n_label'},
    {'image': 'path to q_img'},
    {'text': 'User: It is \nAssistant: '}
]
\end{lstlisting}

%$[$ \texttt{'Is this photo real? Please provide your Answer. You should ONLY output "real" or "fake".'}$, I_1,label_1, I_2,label_2, ... , q_{img}]$.
This structured input is then fed into the \ac{vlm}, which jointly processes visual, textual, and query-specific information. By leveraging this multimodal richness, the model generates a response that effectively integrates retrieved knowledge to improve AI-generated image detection accuracy.

\begin{table*}[!ht]
\caption{Accuracy (ACC) scores of state-of-art detectors and RAVID across 19 test datasets. Best performance is denoted with \textbf{bold}. We report the results of the best RAVID models with different VLMs.}
\label{tab:TrainedOnUniversalFakeDetect}
\begin{adjustbox}{width=\linewidth}
\begin{tabular}{@{}lccccccccccccccccccccc@{}}
\toprule
\multicolumn{1}{c}{\multirow{2}{*}{Methods}} & \multirow{2}{*}{Ref} & \multicolumn{6}{c}{GAN} & \multirow{2}{*}{\begin{tabular}[c]{@{}c@{}}Deep\\ Fakes\end{tabular}} & \multicolumn{2}{c}{Low level} & \multicolumn{2}{c}{Perceptual loss} & \multirow{2}{*}{Guided} & \multicolumn{3}{c}{LDM} & \multicolumn{3}{c}{Glide} & \multirow{2}{*}{Dalle} & \multirow{2}{*}{mAcc} \\ \cmidrule(lr){3-8} \cmidrule(lr){10-13} \cmidrule(lr){15-20}
\multicolumn{1}{c}{} &  & \begin{tabular}[c]{@{}c@{}}Pro-\\ GAN\end{tabular} & \begin{tabular}[c]{@{}c@{}}Cycle-\\ GAN\end{tabular} & \begin{tabular}[c]{@{}c@{}}Big-\\ GAN\end{tabular} & \begin{tabular}[c]{@{}c@{}}Style-\\ GAN\end{tabular} & \begin{tabular}[c]{@{}c@{}}Gau-\\ GAN\end{tabular} & \begin{tabular}[c]{@{}c@{}}Star-\\ GAN\end{tabular} &  & SITD & SAN & CRN & IMLE &  & \begin{tabular}[c]{@{}c@{}}200\\ steps\end{tabular} & \begin{tabular}[c]{@{}c@{}}200\\ w/cfg\end{tabular} & \begin{tabular}[c]{@{}c@{}}100\\ steps\end{tabular} & \begin{tabular}[c]{@{}c@{}}100\\ 27\end{tabular} & \begin{tabular}[c]{@{}c@{}}50\\ 27\end{tabular} & \begin{tabular}[c]{@{}c@{}}100\\ 10\end{tabular} &  &  \\ \cmidrule(r){1-2} \cmidrule(lr){9-9} \cmidrule(lr){14-14} \cmidrule(l){21-22} 

Freq-spec & WIFS2019 & 49.90 & \bf99.90 & 50.50 & 49.90 & 50.30 & 99.70 & 50.10 & 50.00 & 48.00 & 50.60 & 50.10 & 50.90 & 50.40 & 50.40 & 50.30 & 51.70 & 51.40 & 50.40 & 50.00 & 55.45 \\

Co-occurence & Elect. Imag. & 97.70 & 97.70 & 53.75 & 92.50 & 51.10 & 54.70 & 57.10 & 63.06 & 55.85 & 65.65 & 65.80 & 60.50 & 70.70 & 70.55 & 71.00 & 70.25 & 69.60 &  69.90 & 67.55 & 66.86 \\

CNN-Spot & CVPR2020 & 99.99 & 85.20 & 70.20 & 85.70 & 78.95 & 91.70 & 53.47 & 66.67 & 48.69 & 86.31 & 86.26 & 60.07 & 54.03 & 54.96 & 54.14 & 60.78 & 63.80 & 65.66 & 55.58 & 69.58 \\

Patchfor & ECCV2020 & 75.03 & 68.97 & 68.47 & 79.16 & 64.23 & 63.94 & 75.54 & 75.14 & 75.28 & 72.33 & 55.30 & 67.41 & 76.50 & 76.10 & 75.77 & 74.81 & 73.28 & 68.52 & 67.91 & 71.24 \\

F3Net & ECCV2020 & 99.38 & 76.38 & 65.33 & 92.56 & 58.10 & \bf100.00 & 63.48 & 54.17 & 47.26 & 51.47  & 51.47  & 69.20 &  68.15 & 75.35 & 68.80  &  81.65 & 83.25 & 83.05  & 66.30 & 71.33  \\

Bi-LORA & ICASSP2023 & 98.71 & 96.74 & 81.18 & 78.30 & 96.30 & 86.32 & 57.78 & 68.89 & 52.28 & 73.00 & 82.60 & 65.10 & 85.15 & 59.20 & 85.00 & 83.50 & 85.65 & 84.90 & 72.70 & 78.59 \\

LGrad & CVPR2023 & 99.84 & 85.39 & 82.88 & 94.83 & 72.45 & 99.62 & 58.00 & 62.50 & 50.00 & 50.74 & 50.78 & 77.50 & 94.20 & 95.85 & 94.80 & 87.40 & 90.70 & 89.55 & 88.35 & 80.28 \\

UniFD & CVPR2023 & \bf100.00 & 98.50 & 94.50 & 82.00 & 99.50 & 97.00 & 66.60 & 63.00 & 57.50 & 59.50 & 72.00 & 70.03 & 94.19 & 73.76 & 94.36 & 79.07 & 79.85 & 78.14 & 86.78 & 81.38 \\

AntiFakePrompt & CVPR2023 & 99.26 & 96.82 & 87.88 & 80.00 & 98.13 & 83.57 & 60.20 & 70.56 & 53.70 & 79.21 & 79.01 & 73.75 & 89.55 & 64.10 & 89.80 & 93.55 & 93.90 & 92.95 & 80.10 & 82.42 \\

FreqNet & AAAI2024 & 97.90 & 95.84 & 90.45 & \bf97.55 & 90.24 & 93.41 & \bf 97.40 & 88.92 & 59.04 & 71.92 & 67.35 & \bf86.70 & 84.55 & \bf99.58 & 65.56 & 85.69 & \bf97.40 & 88.15 & 59.06 & 85.09 \\

NPR & CVPR2024 & 99.84 & 95.00 & 87.55 & 96.23 & 86.57 & 99.75 & 76.89 & 66.94 & \bf98.63 & 50.00 & 50.00 & 84.55 & 97.65 & 98.00 & 98.20 & \bf96.25 & 97.15 & \bf97.35 & 87.15 & 87.56 \\

FatFormer & CVPR2024 & 99.89 & 99.32 & \bf99.50 & 97.15 & 99.41 & 99.75 & 93.23 & 81.11 & 68.04 & 69.45 & 69.45 & 76.00 & 98.60 & 94.90 & 98.65 & 94.35 & 94.65 & 94.20 & \bf 98.75 & 90.86 \\

RINE & ECCV2024 & 100.00 & 99.30 & 99.60 & 88.90 & \bf 99.80 & 99.50 & 80.60 & 90.60 & 68.30 & 89.20 & 90.60 & 76.10 & 98.30 & 88.20 & 98.60 & 88.90 & 92.60 & 90.70 & 95.00 & 91.31\\

C2P-CLIP$^\star$& Arxiv 2024& 99.71 & 90.69 & 95.28 & \bf99.38 & 95.26 & 96.60 & 89.86 & \bf98.33 & 64.61 & 90.69 & 90.69 & 77.80 & 99.05 & 98.05 & 98.95 & 94.65 & 94.20 & 94.40 & \bf98.80 & 93.00 \\

C2P-CLIP$^\ddagger$ & Arxiv 2024& 99.98 & 97.31 & 99.12 & 96.44 & 99.17 & 99.60 & \bf93.77 & 95.56 & 64.38 & \bf93.29 & 93.29 & 69.10 & \bf99.25 & 97.25 & 99.30 & 95.25 & 95.25 & 96.10 & 98.55 & 93.79 \\

%\rowcolor[HTML]{A5F5DE} RAVID (N=3) & Gemma3 & 92.46 & 89.82 & 76.88 & 82.34 & 90.81 & 70.94 & 59.93 & 71.94 & 63.24 & 56.83 & 56.86 & 58.90 & 86.20 & 78.10 & 85.35 & 83.55 & 83.10 & 84.40 & 79.85 &  76.19 \\
  
\rowcolor[HTML]{D4E8E7} RAVID (N=13) & Gemma3 & 97.34 & 92.73 & 92.92 & 89.68 & 95.55 & 93.42 & 76.11 & 72.22 & 62.33 & 88.62 & 88.37 & 67.25 & 95.60 & 92.45 & 95.55 & 92.50 & 92.90 & 93.30 & 93.60 & 88.02 \\

%\rowcolor[HTML]{F5F5DC} RAVID (N=1) & Qwen-VL &  91.29 & 93.60 & 83.60 & 66.23 & 80.19 &  93.70 & 90.84 & 88.89 & 59.36 & 88.89 & 94.58 & 63.01 & 94.89 & 88.39 & 94.89 & 89.09 & 89.44 & 90.84 & 90.69 & 85.92 \\
 
\rowcolor[HTML]{9FC0FD} RAVID (N=13) & Qwen-VL & 99.96 & 97.84 & 98.70 & 95.24 & 99.28 & 99.82 & 93.36 & 93.61 & 63.01 & 96.47 & 96.46 & 67.80 & 99.30 & 96.75 & 99.40 & 94.05 & 95.00 & 95.70 & 98.40 &  93.69 \\

%\rowcolor[HTML]{E5F5DE} RAVID (N=1) & Openflamingo & 50.00 & 50.00 & 50.00 & 50.00 & 50.00 & 50.00 & 50.08 & 50.00 & 50.00 & 50.00 & 50.00 & 50.00 & 50.00 & 50.00 & 50.00 & 50.00 & 50.00 & 50.00 & 50.00 & 50.00 \\

%\rowcolor[HTML]{E5F5DE} RAVID (N=3) & Openflamingo & 99.95 & 97.84 & 99.25 & 95.94 & 99.38 & 99.85 & 92.64 & 93.61 & 62.33 & 97.55 & 97.59 & 66.95 & 99.35 & 96.35 & 99.35 & 93.10 & 93.25 & 94.40 & 98.40 & 93.53 \\

%\rowcolor[HTML]{E5F5DE} RAVID (N=5) & Openflamingo & 99.95 & 97.58 & 99.28 & 96.24 &  99.30 & 99.82 & 93.19 &  93.89 & 63.24 &  96.26 & 96.25 & 68.10 & 99.20 & 96.45 & 99.30 & 94.00 & 94.35 &  95.10 & 98.30 & 93.67 \\

%\rowcolor[HTML]{E5F5DE} RAVID (N=7) & Openflamingo & 99.96 & 97.46 & 99.15 & 96.24 & 99.32 & 99.80 & 93.30 & 94.72 & 63.93 & 96.04 & 96.03 & 68.30 & 99.25 & 96.65 & 99.35 & 94.00 & 94.70 & 95.30 & 98.15 & 93.77 \\ 

\rowcolor[HTML]{FFC2C2} RAVID (N=13) & Openflamingo & 99.98 & 97.35 & 99.15 & 96.27 & 99.33 & 99.82 & 93.47 & 95.00 & 63.70 & 95.53 & 95.56 & 68.75 & 99.20 & 96.95 & \bf 99.35 & 94.65 & 94.90 & 95.85 & 98.35 & \bf 93.85 \\

%\rowcolor[HTML]{F5F5DC} RAVID (ours) & Compress q=80 & 99.84 & 95.31 & 95.92 & 95.16 & 97.35 & 99.85 & 92.43 & \bf 92.50 & 59.82 & \bf 97.98 & \bf 98.03 & 66.85 & \bf 99.10 & 95.75 & \bf 99.15 & 92.85 & 92.95 & 94.35 & 97.95 & \bf 92.80 \\

% q=20, 99.80 & 95.84 & 94.37 & 93.48 & 97.66 & 99.85 & 92.43 & 92.22 & 61.19 & 97.92 & 97.99 & 67.00 & 99.20 & 96.05 & 99.25 & 92.95 & 93.00 & 94.40 & 98.00 & mAcc = 92.77

% q=10, 99.91 & 97.65 & 98.15 & 95.43 & 99.34 & 99.87 & 92.41 & 92.50 & 61.19 & 97.81 & 97.86 & 67.05 & 99.30 & 96.30 & 99.25 & 92.75 & 93.15 & 94.25 & 98.15 & mAcc = 93.28

%blur 1 & 99.90 & 97.80 & 96.68 & 94.42 & 98.82 & 99.85 & 92.45 & 93.33 & 61.64 & 97.91 & 98.00 & 67.05 & 99.15 & 95.90 & 99.15 & 92.90 & 93.25 & 94.35 & 97.90 & 93.18

%blur 2, 99.95, 97.80, 98.72, 95.67, 99.18, 99.87, 92.41, 93.33, 61.64, 97.95, 98.03, 66.75, 99.15, 96.00, 99.15, 92.70, 92.85, 94.25, 98.00, 93.34

%blur 3, 99.95, 97.92, 99.00, 95.79, 99.27, 99.87, 92.41, 93.89, 62.10, 97.91, 97.99, 66.80, 99.20, 95.95, 99.20, 92.85, 93.05, 94.25, 98.05, mAcc = 93.45

\bottomrule
\end{tabular}

\end{adjustbox}
\begin{flushleft}
($\star$) Trump,Biden. ($\ddagger$)  Deepfake,Camera.
\end{flushleft}
\vspace{-6mm}
\end{table*}

\section{Experimental Results}
\label{sec:experiments}
In this section, we present an extensive evaluation covering multiple aspects, such as datasets, implementation details, and AI-generated image detection performance.
\subsection{Experimental Setup}
\noindent \textbf{Dataset.} To ensure a fair comparison, we utilize the widely recognized UniversalFakeDetect dataset~\cite{ojha2023towards}, which has been extensively used in prior benchmarks. This allows for a direct evaluation of RAVID against state-of-the-art methods, ensuring consistency and robustness in performance assessment. Following the experimental setup introduced by Wang {\it et al.}~\cite{wang2020cnn}, the dataset employs ProGAN as the training set, comprising 20 subsets of generated images. For constructing our vector database, we adopt a 4-class setting (horse, chair, cat, car) as outlined by Tan {\it et al.}~\cite{tan2024c2p}. The test set consists of 19 subsets generated by a diverse range of models, including ProGAN~\cite{karras2018progressive}, StyleGAN~\cite{karras2019style}, BigGAN~\cite{brocklarge}, CycleGAN~\cite{zhu2017unpaired}, StarGAN~\cite{choi2018stargan}, GauGAN~\cite{park2019gaugan}, Deepfake~\cite{rossler2019faceforensics}, CRN~\cite{chen2017photographic}, IMLE~\cite{li2019diverse}, SAN~\cite{dai2019second}, SITD~\cite{chen2018learning}, Guided Diffusion~\cite{dhariwal2021diffusion}, LDM~\cite{rombach2022high}, GLIDE~\cite{nichol2022glide}, and DALLE~\cite{ramesh2021zero}.

%\hl{why 20 it only ProGAN, yes progan contains 20 subsets like car, horse, cat...., one subset for 20 class}

% \begin{itemize}[label=\textbullet]
%     \item \textbf{UniversalFakeDetect~\cite{ojha2023towards}} dataset, follows the setting introduced by (Wang {\it  et al.}, 2020). It uses ProGAN as the training set, which includes 20 subsets of generated images. For training, we adopt a 4-class setting (horse, chair, cat, car) as outlined by Tan {\it et al.}~\cite{tan2024c2p}. The test set consists of 19 subsets from various generative models, including ProGAN~\cite{karras2017progressive}, StyleGAN~\cite{karras2019style}, BigGAN~\cite{karras2019style}, CycleGAN~\cite{karras2019style}, StarGAN ~\cite{choi2018stargan}, GauGAN~\cite{karras2019style}, Deepfake~\cite{rossler2019faceforensics}, CRN~\cite{chen2017photographic}, IMLE~\cite{li2019diverse}, SAN~\cite{dai2019second}, SITD~\cite{chen2018learning}, Guided Diffusion~\cite{dhariwal2021diffusion}, LDM~\cite{rombach2022high}, GLIDE~\cite{nichol2021glide}, and DALLE~\cite{ramesh2021zero}. 
% \end{itemize}

\noindent {\textbf{\\Evaluation Metrics.}} Following the convention of previous detection methods~\cite{keita2024harnessing,tan2024c2p,ojha2023towards,wang2023dire}, we report the accuracy (ACC). We also calculate the mean accuracy across all data subsets to provide a more comprehensive evaluation of overall model performance.\\

\noindent {\textbf{Baselines.}} 
In our study, we fine-tuned AntiFakePrompt~\cite{chang2023antifakeprompt} and Bi-LORA~\cite{keita2025bi,keita2024harnessing}. Moreover, we have chosen the latest and most competitive methods in the field, including Co-occurence~\cite{nataraj2019detecting}, Freq-spec~\cite{zhang2019detecting}, CNN-Spot~\cite{wang2020cnn}, FatchFor~\cite{chai2020makes}, UniFD~\cite{ojha2023towards}, LGrad~\cite{tan2023learning}, F3Net~\cite{qian2020thinking}, FreqNet~\cite{tan2024frequency}, NPR~\cite{tan2024rethinking}, Fatformer~\cite{liu2024forgery}, C2P-CLIP~\cite{tan2024c2p}, RINE~\cite{koutlis2024leveraging}, for all those models, we report the results presented in~\cite{tan2024c2p}. Finally, for RINE, we report results from its paper~\cite{koutlis2024leveraging}.\\

%\hl{not clar here why these two: cuz they do not have been trained on this data, so for fair comparison we should trained them also on the same dataset}

%C2P-CLIP~\cite{tan2024c2p}

\noindent\textbf{Implementation Details.} To fine-tune CLIP ViT-L/14, we use Adam optimizer with an initial learning rate of $4 \times 10^{-4}$, a batch size of $128$, and train for 1 epoch. We apply LoRA layers to the $q\_proj$, $k\_proj$, and $v\_proj$ layers using the Parameter-Efficient Fine-Tuning (PEFT) library. The LoRA hyper-parameters are set as follows: $lora\_r$ = 6, $lora\_alpha$ = 6, and $lora\_dropout$ = 0.8. For the vector database, we use Milvus locally via Docker. On the other hand, for image-augmented response generation, we use Openflamingo~\cite{awadalla2023openflamingo}.  

%In our experiments, we use the PyTorch deep learning framework on an AWS Linux computer equipped with a 24GB NVIDIA RTX A10G. 

\subsection{Comparison with the State-of-the-art}

Table~\ref{tab:TrainedOnUniversalFakeDetect} presents the mean accuracy (mAcc) scores for cross-generator detection on the UniversalFakeDetect dataset, which includes 19 different generative models spanning GANs, Deepfakes, low-level vision models, perceptual loss models, and diffusion models. RAVID achieves 93.85\% mAcc, outperforming 15 state-of-the-art methods and demonstrating strong generalization across diverse image synthesis techniques.

%, and the comparative trends is shown in Figue~\ref{fig:acc}.

The baseline methods, such as Freq-spec and CNN-Spot, rely on traditional frequency analysis and convolutional neural networks for detecting AI-generated images. In contrast, RAVID adopts a novel approach to enhance feature representation, ensuring better generalization across various generators. Compared to UniFD, a recent state-of-the-art method, RAVID improves the mAcc by 12.47\%, highlighting the effectiveness of our approach. Additionally, RAVID demonstrates competitive performance with the latest methods, RINE and C2P-CLIP, achieving a mere 1.48\% and 0.16\% mAcc gap, respectively. While RINE utilizes advanced feature extraction and fusion techniques, which increase computational complexity, C2P-CLIP embeds category-specific concepts into CLIP's image encoder. Meanwhile, our method strikes a balance between performance and efficiency, making it more suitable for real-world applications.
%\hl{how you claim this: by empirical results from  experiments comparing RAVID with multiple baseline methods}

We also analyze detection performance across different generator categories. For Perceptual Loss models, RAVID achieves an impressive detection accuracy of 95.56\%, while for Low-level Vision models, it achieves 95.00\%, demonstrating competitive performance with all other methods in these categories. This demonstrates RAVID's ability to effectively handle challenging generators without requiring specialized architectural modifications or additional computational resources. 

%Figure~\ref{fig:acc} provides insight into the trade-off between accuracy and model complexity, where RAVID demonstrates competitive performance while maintaining efficiency.

\begin{table*}[!ht]
\caption{In-context learning without RAG. Instead of retrieving relevant images to the query image, we randomly select $N$ images from the 4-class setting ProGAN training set. Best performance is denoted with \textbf{bold}.}
\label{tab:RAVIDInContextLearning}
\begin{adjustbox}{width=\linewidth}
\begin{tabular}{@{}ccccccccccccccccccccccc@{}}
\toprule
\multirow{2}{*}{Methods} & \multirow{2}{*}{VLMs} & \multirow{2}{*}{N Shots} & \multicolumn{6}{c}{GAN} & \multirow{2}{*}{\begin{tabular}[c]{@{}c@{}}Deep\\ Fakes\end{tabular}} & \multicolumn{2}{c}{Low level} & \multicolumn{2}{c}{Perceptual loss} & \multirow{2}{*}{Guided} & \multicolumn{3}{c}{LDM} & \multicolumn{3}{c}{Glide} & \multirow{2}{*}{Dalle} & \multirow{2}{*}{mAcc} \\ \cmidrule(lr){4-9} \cmidrule(lr){11-14} \cmidrule(lr){16-21}
 &  &  & \begin{tabular}[c]{@{}c@{}}Pro-\\ GAN\end{tabular} & \begin{tabular}[c]{@{}c@{}}Cycle-\\ GAN\end{tabular} & \begin{tabular}[c]{@{}c@{}}Big-\\ GAN\end{tabular} & \begin{tabular}[c]{@{}c@{}}Style-\\ GAN\end{tabular} & \begin{tabular}[c]{@{}c@{}}Gau-\\ GAN\end{tabular} & \begin{tabular}[c]{@{}c@{}}Star-\\ GAN\end{tabular} &  & SITD & SAN & CRN & IMLE &  & \begin{tabular}[c]{@{}c@{}}200\\ steps\end{tabular} & \begin{tabular}[c]{@{}c@{}}200\\ w/cfg\end{tabular} & \begin{tabular}[c]{@{}c@{}}100\\ steps\end{tabular} & \begin{tabular}[c]{@{}c@{}}100\\ 27\end{tabular} & \begin{tabular}[c]{@{}c@{}}50\\ 27\end{tabular} & \begin{tabular}[c]{@{}c@{}}100\\ 10\end{tabular} &  &  \\ \cmidrule(r){1-3} \cmidrule(lr){10-10} \cmidrule(lr){15-15} \cmidrule(l){22-23}
 
%\multicolumn{1}{l}{RAVID W/ RAG CLIP} & QWen-VL & 3 & 98.17 & 92.51 & 85.77 & 74.64 & 93.87 & 98.25 & 66.88 & 72.22 & 57.31 & 61.85 & 72.46 & 68.88 & 88.34 & 66.88 & 89.14 & 84.89 & 85.29 & 86.14 & 79.39 & 80.15 \\
 
% \multicolumn{1}{l}{RAVID W/O RAG} & QWen-VL & 1 & 58.39 & 61.02 & 53.83 & 49.88 & 56.13 & 50.18 & 51.47 & 52.66 & \bf 61.42 & 46.01 & 47.13 & \bf 51.91 & 53.57 & 53.15 & 54.18 & 55.62 & 54.48 & 55.54 & 51.92 & 53.60 \\

% \multicolumn{1}{l}{RAVID W/O RAG} & QWen-VL & 3 & \bf 65.80 & \bf 65.63 & \bf 55.38 & \bf 52.25 & \bf 59.27 & \bf 53.01 & \bf 52.06 & \bf 56.39 & 55.71 & \bf 51.49 & \bf 56.53 & 51.65 & \bf 61.82 & \bf 53.33 & \bf 63.59 & \bf 60.65 & \bf 59.52 & \bf 59.24 & \bf 53.68 & \bf 57.28 \\

% \rowcolor[HTML]{FFC2C2} \multicolumn{1}{l}{RAVID W/ RAG} & Openflamingo & 1 &  &  &  &  &  &  &  &  &  &  &  &  &  &  &  &  &  &  &  &  \\

\multicolumn{1}{l}{RAVID W/ RAG} & Openflamingo & 3 & 99.95 & 97.84 & 99.25 & 95.94 & 99.38 & 99.85 & 92.64 & 93.61 & 62.33 & 97.55 & 97.59 & 66.95 & 99.35 & 96.35 & 99.35 & 93.10 & 93.25 & 94.40 & 98.40 & 93.53 \\

\multicolumn{1}{l}{RAVID W/ RAG} & Openflamingo & 13 & 99.98 & 97.35 & 99.15 & 96.27 & 99.33 & 99.82 & 93.47 & 95.00 & 63.70 & 95.53 & 95.56 & 68.75 & 99.20 & 96.95 & 99.35 & 94.65 & 94.90 & 95.85 & 98.35 & 93.85 \\ 

\rowcolor[HTML]{FFC2C2} \multicolumn{1}{l}{RAVID W/O RAG} & Openflamingo & 3 &49.49 & 50.53 & 51.00 & 49.99 & 49.64 & 50.88 & 50.53 & 49.17 & 45.89 & 49.89 & 49.87 & 51.40 & 50.65 & 50.40 & 50.35 & 50.30 & 50.45 & 50.95 & 50.45 & 50.10 \\

\rowcolor[HTML]{FFC2C2} \multicolumn{1}{l}{RAVID W/O RAG} & Openflamingo & 13 & 49.38 & 51.59 & 50.78 & 49.67 & 49.80 & 50.55 & 51.14 & 50.00 & 47.72 & 50.65 & 49.85 & 48.50 & 49.25 & 49.85 & 48.90 & 49.70 & 50.50 & 51.55 & 48.80 & 49.90 \\

% \rowcolor[HTML]{FFC2C2} \multicolumn{1}{l}{RAVID W/ RAG} & Openflamingo & 5 &  &  &  &  &  &  &  &  &  &  &  &  &  &  &  &  &  &  &  &  \\

% \rowcolor[HTML]{FFC2C2} \multicolumn{1}{l}{RAVID W/ RAG} & Openflamingo & 7 &  &  &  &  &  &  &  &  &  &  &  &  &  &  &  &  &  &  &  &  \\

\bottomrule
\end{tabular}
\end{adjustbox}
\end{table*}

\begin{table*}[!ht]
\caption{Comparison of image retrieval performance in RAVID: Fine-Tuned vs. Non-Fine-Tuned CLIP as an image embedding model for retrieving relevant images in the context of AI-generated images. The notation (\textbf{*}) denotes RAVID using a fine-tuned CLIP model as the image embedding method.}
\label{tab:RAVIDICLIPnoFune}
\begin{adjustbox}{width=\linewidth}
\begin{tabular}{@{}ccccccccccccccccccccccc@{}}
\toprule
\multirow{2}{*}{Methods} & \multirow{2}{*}{VLMs} & \multirow{2}{*}{N Shots} & \multicolumn{6}{c}{GAN} & \multirow{2}{*}{\begin{tabular}[c]{@{}c@{}}Deep\\ Fakes\end{tabular}} & \multicolumn{2}{c}{Low level} & \multicolumn{2}{c}{Perceptual loss} & \multirow{2}{*}{Guided} & \multicolumn{3}{c}{LDM} & \multicolumn{3}{c}{Glide} & \multirow{2}{*}{Dalle} & \multirow{2}{*}{mAcc} \\ \cmidrule(lr){4-9} \cmidrule(lr){11-14} \cmidrule(lr){16-21}
 &  &  & \begin{tabular}[c]{@{}c@{}}Pro-\\ GAN\end{tabular} & \begin{tabular}[c]{@{}c@{}}Cycle-\\ GAN\end{tabular} & \begin{tabular}[c]{@{}c@{}}Big-\\ GAN\end{tabular} & \begin{tabular}[c]{@{}c@{}}Style-\\ GAN\end{tabular} & \begin{tabular}[c]{@{}c@{}}Gau-\\ GAN\end{tabular} & \begin{tabular}[c]{@{}c@{}}Star-\\ GAN\end{tabular} &  & SITD & SAN & CRN & IMLE &  & \begin{tabular}[c]{@{}c@{}}200\\ steps\end{tabular} & \begin{tabular}[c]{@{}c@{}}200\\ w/cfg\end{tabular} & \begin{tabular}[c]{@{}c@{}}100\\ steps\end{tabular} & \begin{tabular}[c]{@{}c@{}}100\\ 27\end{tabular} & \begin{tabular}[c]{@{}c@{}}50\\ 27\end{tabular} & \begin{tabular}[c]{@{}c@{}}100\\ 10\end{tabular} &  &  \\ \cmidrule(r){1-3} \cmidrule(lr){10-10} \cmidrule(lr){15-15} \cmidrule(l){22-23}

% \rowcolor[HTML]{F5F5DC} \multicolumn{1}{l}{RAVID W/ RAG (*)} & Qwen-VL & 1 & 91.29 & 93.60 & 83.60 & 66.23 & 80.19 &  93.70 & 90.84 & 88.89 & 59.36 & 88.89 & 94.58 & 63.01 & 94.89 & 88.39 & 94.89 & 89.09 & 89.44 & 90.84 & 90.69 & 85.92 \\
 
% \rowcolor[HTML]{F5F5DC} \multicolumn{1}{l}{RAVID W/ RAG (*)} & Qwen-VL & 3 & 99.49 & 97.27 & 94.77 & 93.16 & 96.11 & 99.77 & 92.45 & \bf 93.33 & 62.33 & {\bf 97.91} & {\bf 
%  98.00} & 66.60 & \bf 99.00 & 95.55 & \bf 99.10 & 92.90 & 93.05 & 94.55 & 97.75 & \bf 92.79 \\

% \multicolumn{1}{l}{RAVID W/ RAG } & QWen-VL & 1 & 90.54 & 87.62 & 76.38 & 63.12 & 78.06 & 87.94 &  65.96 & 66.85 & 56.39 & 57.21 & 68.27 & 66.15 & 83.41 & 66.23 & 85.25 & 83.36 & 83.71 & 83.91 & 75.13 & 75.03 \\

% \multicolumn{1}{l}{RAVID W/ RAG} & QWen-VL & 3 & 98.17 & 92.51 & 85.77 & 74.64 & 93.87 & 98.25 & 66.88 & 72.22 & 57.31 & 61.85 & 72.46 & 68.88 & 88.34 & 66.88 & 89.14 & 84.89 & 85.29 & 86.14 & 79.39 & 80.15 \\ \midrule

\multicolumn{1}{l}{RAVID W/ RAG (*)} & Openflamingo & 1 & 50.00 & 50.00 & 50.00 & 50.00 & 50.00 & 50.00 & 50.08 & 50.00 & 50.00 & 50.00 & 50.00 & 50.00 & 50.00 & 50.00 & 50.00 & 50.00 & 50.00 & 50.00 & 50.00 & 50.00 \\

\multicolumn{1}{l}{RAVID W/ RAG (*)} & Openflamingo  & 3 & 99.95 & 97.84 & 99.25 & 95.94 & 99.38 & 99.85 & 92.64 & 93.61 & 62.33 & 97.55 & 97.59 & 66.95 & 99.35 & 96.35 & 99.35 & 93.10 & 93.25 & 94.40 & 98.40 & 93.53 \\

\multicolumn{1}{l}{RAVID W/ RAG (*)} & Openflamingo & 5 & 99.95 & 97.58 & 99.28 & 96.24 &  99.30 & 99.82 & 93.19 &  93.89 & 63.24 &  96.26 & 96.25 & 68.10 & 99.20 & 96.45 & 99.30 & 94.00 & 94.35 &  95.10 & 98.30 & 93.67 \\

\multicolumn{1}{l}{RAVID W/ RAG (*)} & Openflamingo & 7 & 99.96 & 97.46 & 99.15 & 96.24 & 99.32 & 99.80 & 93.30 & 94.72 & 63.93 & 96.04 & 96.03 & 68.30 & 99.25 & 96.65 & 99.35 & 94.00 & 94.70 & 95.30 & 98.15 & 93.77 \\ 

\multicolumn{1}{l}{RAVID W/ RAG (*)} & Openflamingo  & 13 & 99.98 & 97.35 & 99.15 & 96.27 & 99.33 & 99.82 & 93.47 & 95.00 & 63.70 & 95.53 & 95.56 & 68.75 & 99.20 & 96.95 & 99.35 & 94.65 & 94.90 & 95.85 & 98.35 & 93.85 \\

\rowcolor[HTML]{FFC2C2} \multicolumn{1}{l}{RAVID W/ RAG} & Openflamingo & 1 & 50.00 & 50.00 & 50.00 & 50.00 & 50.00 & 50.00 & 50.08 & 50.00 & 50.00 & 50.00 & 50.00 & 50.00 & 50.00 & 50.00 & 50.00 & 50.00 & 50.00 & 50.00 & 50.00 & 50.00 \\

\rowcolor[HTML]{FFC2C2} \multicolumn{1}{l}{RAVID W/ RAG} & Openflamingo & 3 & 97.76 & 89.06 & 86.93 & 76.91 & 95.34 & 98.32 & 70.53 & 70.83 & 56.85 & 67.11 & 76.90 & 69.30 & 85.65 & 67.30 & 85.85 & 80.20 & 81.05 & 82.00 & 78.15 & 79.79 \\

\rowcolor[HTML]{FFC2C2} \multicolumn{1}{l}{RAVID W/ RAG} & Openflamingo & 5 & 98.09 & 89.74 & 88.20 & 77.59 & 95.30 & 98.62 & 73.08 & 65.28 & 59.13 & 71.01 & 79.20 & 71.55 & 86.90 & 69.90 & 88.25 & 84.90 & 85.10 & 84.50 & 80.80 & 81.43 \\

\rowcolor[HTML]{FFC2C2} \multicolumn{1}{l}{RAVID W/ RAG} & Openflamingo & 7 & 98.16 & 89.86 & 88.00 & 78.13 & 95.06 & 98.57 & 74.04 & 65.83 & 61.42 & 69.12 & 77.94 & 71.75 & 87.60 & 71.75 & 88.35 & 86.10 & 86.70 & 86.00 & 82.35 & 81.93 \\

\rowcolor[HTML]{FFC2C2} \multicolumn{1}{l}{RAVID W/ RAG} & Openflamingo & 13 & 98.23 & 88.72 & 88.13 & 77.61 & 94.68 & 98.37 & 75.30 & 69.17 & 58.90 & 68.10 & 76.32 & 71.75 & 87.70 & 71.40 & 88.40 & 86.30 & 87.40 & 86.70 & 82.45 & 81.88 \\

\bottomrule
\end{tabular}
\end{adjustbox}
\end{table*}

\subsection{Impact of Retrieved Image Count on Detection Performance}

To evaluate the impact of the number of retrieved images on RAVID's AI-generated image detection performance, we conducted experiments with varying retrieval settings. Specifically, we compared performance when retrieving 1 ($N=1$), 3 ($N=3$), 5 ($N=5$), 7 ($N=7$), and 13 ($N=13$) images, utilizing the Openflamingo vision-language model. The results, presented in Table~\ref{tab:diffVLMs}, show a substantial improvement in detection accuracy as the number of retrieved images increases. With one retrieved image, the model achieved an average accuracy (mAcc) of 50.00\%, whereas increasing the retrieval count to $N=3$ led to a significant boost to 93.53\%. This improvement was consistent across various generative models, such as ProGAN (50.00\% → 99.95\%), StarGAN (50.00\% → 99.85\%), and DeepFakes (50.08\% → 92.64\%). As the number of retrieved images increases to $N=$5, $N=$7 and $N=$13, mAcc increases progressively to 93.67\%, 93.77\% and 93.85\%, respectively. These findings indicate that incorporating additional retrieved images provides richer contextual information, thereby improving the model’s generalization across diverse AI-generated image models.

Figure~\ref{fig:Nshot} RAVID's detection performance across different categories of AI-generated images, with varying numbers of retrieved images (shots). As the number of retrieved images increases from 1 to 3, a substantial improvement in detection accuracy is observed for several generative models. This indicates that incorporating more contextual information enhances RAVID's ability to identify AI-generated content.  However, the performance gains become minimal as the number of retrieved images exceeds  3. This highlights how additional context enhances RAVID's robustness in distinguishing AI-generated images. Overall, these results underscore the value of leveraging retrieval-augmented generation for improved detection performance.

\begin{figure*}[!ht]
    \centering
    \includegraphics[width=\linewidth]{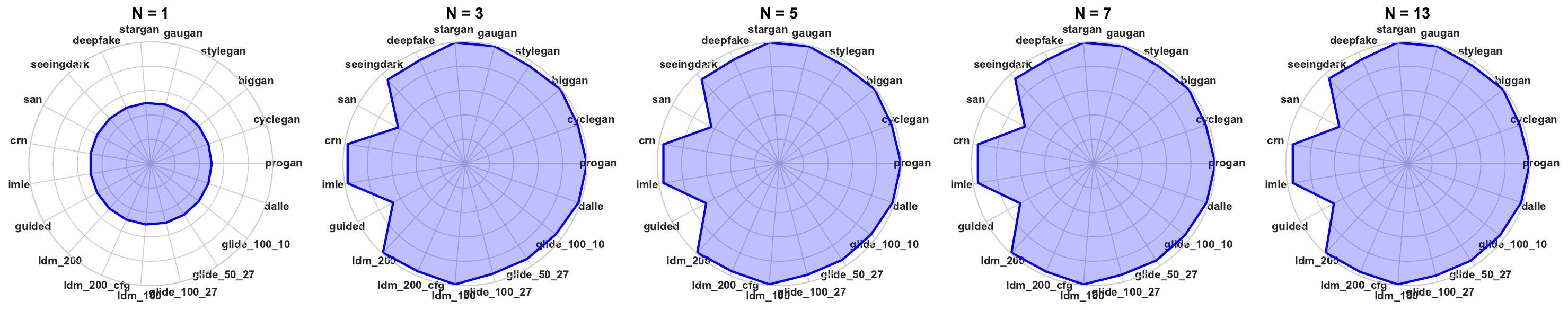}
    \caption{Performance comparison of RAVID W/ CLIP fine-tuned across different N-shot settings.}
    \label{fig:Nshot}
\end{figure*}

\begin{figure}[!ht]
    \centering
    \includegraphics[width=0.8\linewidth]{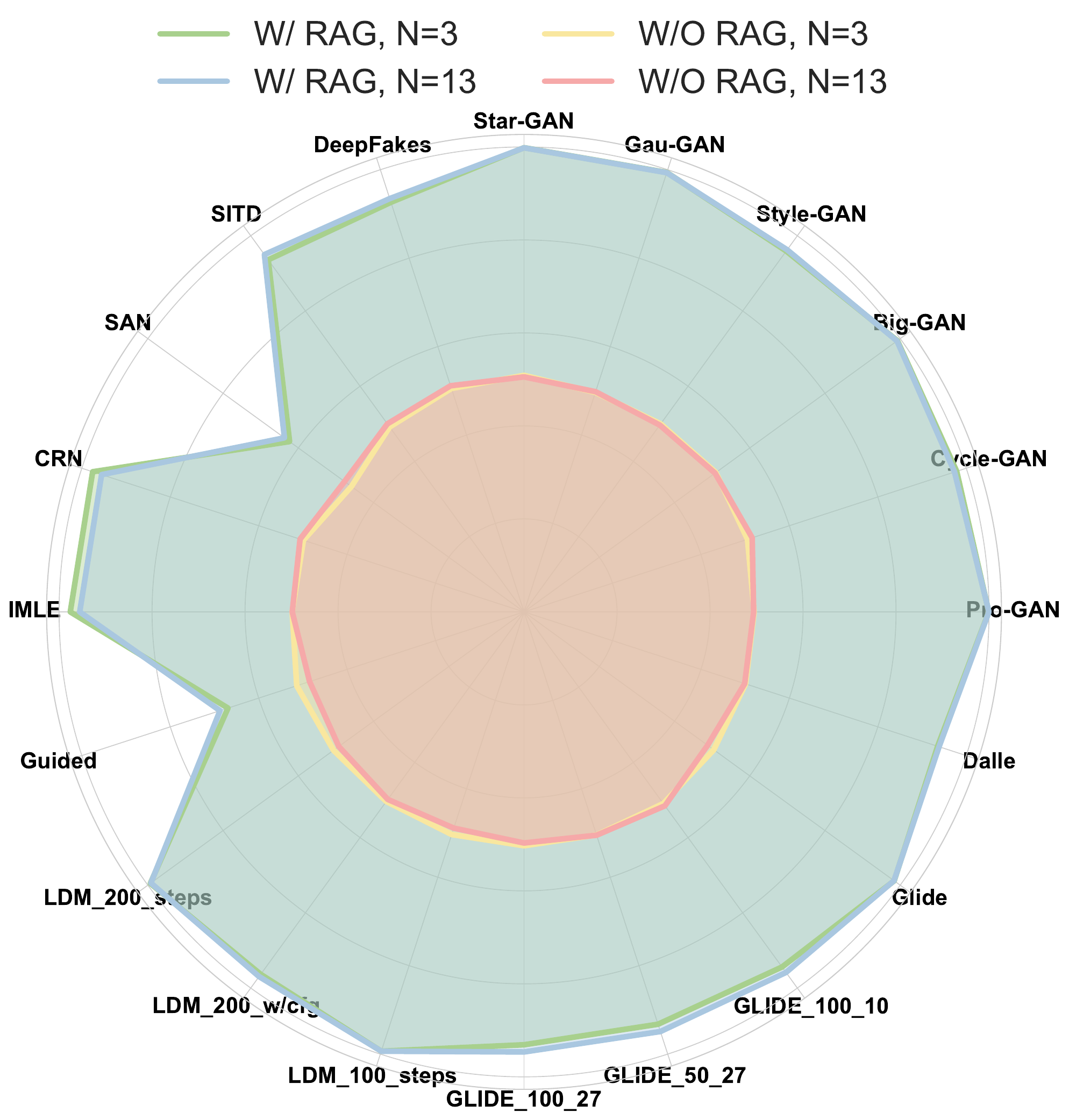}
    \caption{Performance comparison of RAVID W/ RAG and W/O RAG across different N-shot settings. The accuracy shift demonstrates the positive impact of retrieval, with higher performance in the W/RAG setup.}
    \label{fig:Nshot}
\end{figure}

\subsection{Impact of Using RAG for Retrieval in RAVID}

To evaluate the impact of dynamically retrieving relevant images in the RAVID approach, we conducted an experiment where the context provided to the \ac{vlm} was formed by randomly selecting images, instead of using the RAG retrieval mechanism. This experiment allowed us to assess the significance of the retrieval process in improving detection accuracy. In this setup, rather than retrieving relevant images related to the query image, we randomly selected $N$ images from the 4-class setting ProGAN training set. These randomly chosen images, along with their corresponding labels, were used as a context for the detection task. This mimicked the in-context learning strategy used in \ac{rag}, but without its retrieval component. We maintained the same configurations as in RAVID, varying the number of selected images (shots). In the 3-shot setup, three randomly selected image was provided as context, while in the 13-shot setup, thirteen images were used as context.

The results in Table~\ref{tab:RAVIDInContextLearning} show the detection accuracy ($mAcc$) across a range of generative models. For the 3-shot setup, RAVID W/O RAG achieved an average accuracy of 50.10\%, whereas in the 13-shot setup, the accuracy slightly declined to 49.90\%. While these results indicate that providing more context does not improve performance, they still fall behind the accuracy achieved when using \ac{rag}, where the retrieved context is more relevant to the query image. This experiment highlights the importance of relevant context in AI-generated image detection. When the model relies on randomly selected images, the context lacks meaningful relevance to the query, limiting its ability to make accurate predictions, especially with complex generative models. In contrast, the ability of \ac{rag} to retrieve relevant images substantially boosts the model’s performance, emphasizing the critical role of relevant context in improving detection accuracy and generalization. This investigation also underscores the need for an image embedding model for retrieval that is sensitive to the subtle characteristics of AI-generated images, as opposed to one that focuses on general cues, which are less useful for this task.

To quantify the impact of retrieval, we analyze, the accuracy gap between RAG-based and non-RAG-based setups. This difference, mAcc(W/ RAG) – mAcc(W/O RAG), directly measures the benefit of retrieving relevant images. Specifically, when RAVID uses \ac{rag} with three retrieval shot, the mean accuracy ($mAcc$) improves from 50.10\% (W/O RAG) to 93.53\% (W/ RAG), representing a significant increase of 43.43\%.  And if we increase the retrieval shot to 13, the average accuracy rises from 49.90\%(W/O RAG ) to 93.85\% (W/ RAG), representing an increase of 43.95\%.  As illustrated in Figure~\ref{fig:Nshot}, this significant improvement is consistently positive, confirming that retrieval improves AI-generated image detection. The consistently higher accuracy of W/ RAG over W/O RAG underscores the importance of meaningful context. Without retrieval, the model relies on irrelevant information, leading to weaker feature alignment and lower accuracy. A well-designed retrieval mechanism ensures that retrieved images share key characteristics with the query, enhancing generalization and improving detection performance.

\begin{figure*}[!th]
    \centering
    \includegraphics[width=0.9\linewidth]{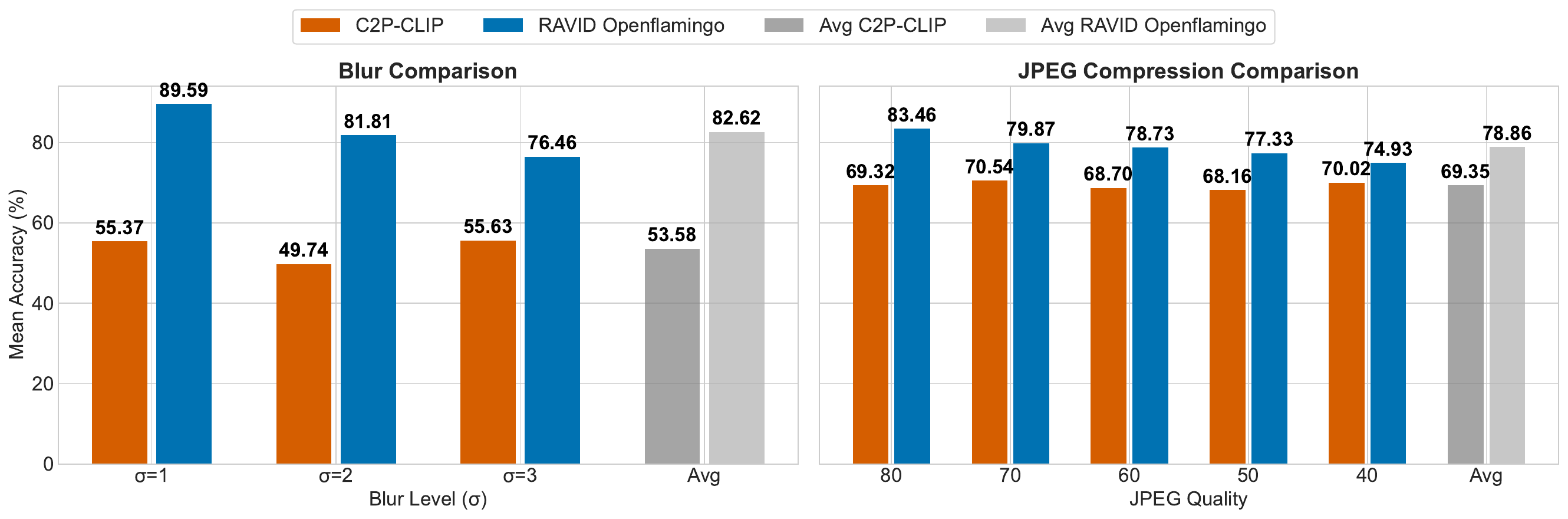}
    \caption{RAVID's robustness under gaussian blur and JPEG compression, common real-world degradations affecting AI-generated image detection. }
    \label{fig:deg}
\end{figure*}

\begin{table*}[!ht]
\caption{Performance (ACC) after applying common image degradations.}
\label{tab:RAVIDICLIPDegradation}
\begin{adjustbox}{width=\linewidth}
\begin{tabular}{@{}lcccccccccccccccccccccc@{}}
\toprule
\multirow{2}{*}{Methods} & \multirow{2}{*}{VLMs} & \multirow{2}{*}{Degradation} & \multicolumn{6}{c}{GAN} & \multirow{2}{*}{\begin{tabular}[c]{@{}c@{}}Deep\\ Fakes\end{tabular}} & \multicolumn{2}{c}{Low level} & \multicolumn{2}{c}{Perceptual loss} & \multirow{2}{*}{Guided} & \multicolumn{3}{c}{LDM} & \multicolumn{3}{c}{Glide} & \multirow{2}{*}{Dalle} & \multirow{2}{*}{mAcc} \\ \cmidrule(lr){4-9} \cmidrule(lr){11-14} \cmidrule(lr){16-21}
 &  &  & \begin{tabular}[c]{@{}c@{}}Pro-\\ GAN\end{tabular} & \begin{tabular}[c]{@{}c@{}}Cycle-\\ GAN\end{tabular} & \begin{tabular}[c]{@{}c@{}}Big-\\ GAN\end{tabular} & \begin{tabular}[c]{@{}c@{}}Style-\\ GAN\end{tabular} & \begin{tabular}[c]{@{}c@{}}Gau-\\ GAN\end{tabular} & \begin{tabular}[c]{@{}c@{}}Star-\\ GAN\end{tabular} &  & SITD & SAN & CRN & IMLE &  & \begin{tabular}[c]{@{}c@{}}200\\ steps\end{tabular} & \begin{tabular}[c]{@{}c@{}}200\\ w/cfg\end{tabular} & \begin{tabular}[c]{@{}c@{}}100\\ steps\end{tabular} & \begin{tabular}[c]{@{}c@{}}100\\ 27\end{tabular} & \begin{tabular}[c]{@{}c@{}}50\\ 27\end{tabular} & \begin{tabular}[c]{@{}c@{}}100\\ 10\end{tabular} &  &  \\ \cmidrule(r){1-3} \cmidrule(lr){10-10} \cmidrule(lr){15-15} \cmidrule(l){22-23}

C2P-CLIP & - & Blur \(\sigma =1\) & 96.10 & 90.31 & 97.02 & 97.00 & 95.75 & 96.80 & 93.43 & 95.56 & 57.08 & 68.84 & 68.84 & 47.90 & 01.20 & 06.60 & 01.60 & 12.50 & 13.30 & 10.60 & 01.60 & 55.37 \\ 

C2P-CLIP & - & Blur \(\sigma =2\) & 72.20 & 85.24 & 87.35 & 79.45 & 90.08 & 86.47 & 80.33 & 95.56 & 51.60 & 60.90 & 61.17 & 45.40 & 01.20 & 06.60 & 01.60 & 12.50 & 13.30 & 10.60 & 03.50 & 49.74 \\

C2P-CLIP & - & Blur \(\sigma =3\) & 77.20 & 84.18 & 77.03 & 62.87 & 87.19 & 88.64 & 75.52 & 95.00 & 49.54 & 56.39 & 56.39 & 47.80 & 13.50 & 35.30 & 12.20 & 42.00 & 40.40 & 42.20 & 13.70 & 55.63 \\ \midrule

% \rowcolor[HTML]{9FC0FD} {RAVID (N=3)} & QWen-VL & Blur \(\sigma =1\) & 97.69 & 92.01 & 95.23 & 95.46 & 96.68 & 98.10 & 92.75 & 93.06 & 56.62 & 83.04 & 83.05 & 71.95 & 97.85 & 93.65 & 97.85 & 90.40 & 89.60 & 91.25 & 97.30 & 90.19 \\

% \rowcolor[HTML]{9FC0FD} {RAVID (N=3)} & QWen-VL & blur \(\sigma =2\) & 78.03 & 87.59 & 88.85 & 82.13 & 93.07 & 91.42 & 82.16 & 93.33 & 50.46 & 72.23 & 75.13 & 71.70 & 94.90 & 87.20 & 95.05 & 78.60 & 80.45 & 79.20 & 95.05 & 82.98 \\

% \rowcolor[HTML]{9FC0FD} {RAVID (N=3)} & QWen-VL & Blur \(\sigma =3\) & 82.29 & 86.71 & 76.20 & 63.73 & 89.82 & 92.55 & 75.45 & 93.33 & 49.54 & 67.98 & 68.83 & 67.70 & 86.60 & 73.35 & 86.85 & 70.20 & 70.60 & 69.50 & 86.65 & 76.73 \\ \midrule

\rowcolor[HTML]{FFC2C2} {RAVID (N=13)} & Openflamingo & Blur \(\sigma =1\)  &  96.50 & 90.46 & 97.28 & 96.90 & 96.17 & 97.47 & 93.23 & 95.00 & 56.85 & 74.12 & 74.11 & 73.35 & 97.40 & 94.30 & 97.25 & 91.25 & 90.90 & 92.75 & 97.00 & 89.59 \\

\rowcolor[HTML]{FFC2C2} {RAVID (N=13)} & Openflamingo & Blur \(\sigma =2\)  &  73.73 & 85.47 & 87.83 & 79.99 & 91.12 & 88.44 & 81.35 & 95.00 & 51.37 & 65.37 & 66.07 & 73.40 & 94.05 & 88.65 & 94.30 & 80.80 & 82.05 & 81.10 & 94.30 & 81.81 \\

\rowcolor[HTML]{FFC2C2} {RAVID (N=13)} & Openflamingo & Blur \(\sigma =3\)  &  78.51 & 84.56 & 76.78 & 63.35 & 88.10 & 90.17 & 76.47 & 95.00 & 49.54 & 59.97 & 60.15 & 70.30 & 87.80 & 75.70 & 88.25 & 73.30 & 74.25 & 73.20 & 87.25 & 76.46 \\ \midrule

C2P-CLIP & - & Jpeq \(q= 80\) & 95.80 & 94.93 & 92.92 & 75.60 & 96.85 & 95.02 & 85.57 & 94.72 & 55.94 & 92.53 & 92.42 & 64.80 & 14.30 & 53.10 & 16.80 & 58.40 & 63.40 & 56.80 & 17.10 & 69.32\\

C2P-CLIP & - & Jpeq \(q= 70\) & 94.49 & 94.44 & 87.10 & 65.30 & 95.18 & 92.72 & 84.90 & 93.89 & 54.79 & 86.45 & 88.09 & 70.30 & 24.10 & 67.70 & 27.90 & 61.70 & 64.10 & 56.70 & 30.50 & 70.54 \\

C2P-CLIP & - & Jpeq \(q= 60\) & 94.59 & 94.40 & 81.80 & 62.30 & 95.57 & 89.62 & 82.76 & 90.56 & 53.65 & 80.00 & 80.05 & 65.90 & 23.20 & 69.70 & 26.10 & 58.00 & 60.50 & 54.50 & 42.10 & 68.70 \\

C2P-CLIP & - & Jpeq \(q= 50\) & 93.79 & 93.26 & 80.55 & 60.17 & 94.69 & 92.12 & 78.74 & 88.06 & 52.97 & 76.64 & 73.89 & 63.40 & 25.70 & 71.40 & 25.20 & 59.90 & 62.60 & 56.50 & 45.40 & 68.16 \\

C2P-CLIP & - & Jpeq \(q= 40\) & 93.23 & 91.79 & 75.55 & 57.20 & 93.22 & 92.62 & 75.10 & 81.39 & 52.74 & 77.96 & 75.56 & 71.50 & 35.10 & 77.80 & 34.60 & 64.90 & 65.10 & 62.20 & 52.80 & 70.02 \\ \midrule

% \rowcolor[HTML]{9FC0FD} {RAVID (N=3)} & QWen-VL & Jpeq \(q= 80\) & 95.89 & 94.44 & 89.54 & 73.49 & 95.30 & 97.02 & 81.46 & 90.83 & 54.11 & 86.46 & 89.10 & 64.05 & 90.00 & 68.35 & 89.25 & 67.00 & 65.35 & 68.20 & 88.00 & 81.47 \\

% \rowcolor[HTML]{9FC0FD} {RAVID (N=3)} & QWen-VL & Jpeq \(q= 70\) & 93.88 & 93.79 & 83.60 & 63.60 & 93.37 & 94.95 & 81.96 & 90.28 & 52.97 & 73.57 & 77.05 & 61.80 & 84.80 & 62.40 & 83.00 & 66.35 & 64.30 & 68.15 & 80.60 & 77.39 \\

% \rowcolor[HTML]{9FC0FD} {RAVID (N=3)} & QWen-VL & Jpeq \(q= 60\) & 94.12 & 93.64 & 78.44 & 60.47 & 93.63 & 94.02 & 80.30 & 86.11 & 52.51 & 78.82 & 80.00 & 63.45 & 85.45 & 61.53 & 83.95 & 67.65 & 66.15 & 69.15 & 75.05 & 77.08 \\

% \rowcolor[HTML]{9FC0FD} {RAVID (N=3)} & QWen-VL & Jpeq \(q= 50\) & 93.77 & 93.34 & 77.08 & 58.36 & 92.79 & 95.57 & 75.62 & 84.17 & 51.14 & 76.75 & 75.42 & 64.73 & 84.30 & 60.75 & 84.25 & 66.05 & 65.10 & 67.00 & 72.49 & 75.72 \\

% \rowcolor[HTML]{9FC0FD} {RAVID (N=3)} & QWen-VL & Jpeq \(q= 40\) & 92.20 & 90.46 & 71.72 & 55.75 & 90.88 & 93.20 & 71.30 & 78.06 & 51.37 & 68.29 & 69.07 & 61.90 & 79.90 & 58.95 & 79.55 & 64.30 & 64.05 & 65.35 & 69.45 & 72.41 \\  \midrule

\rowcolor[HTML]{FFC2C2} {RAVID (N=13)} & Openflamingo & Jpeq \(q= 80\) & 95.90 & 95.31 & 92.68 & 75.07 & 97.00 & 96.00 & 84.14 & 93.61 & 55.71 & 92.96 & 93.32 & 66.15 & 91.20 & 70.95 & 89.90 & 69.20 & 66.65 & 70.15 & 89.80 & 83.46 \\

\rowcolor[HTML]{FFC2C2} {RAVID (N=13)} & Openflamingo & Jpeq \(q= 70\) &  94.24 & 94.55 & 86.58 & 65.05 & 95.09 & 93.85 & 84.16 & 93.33 & 54.57 & 83.85 & 86.38 & 63.65 & 86.20 & 64.60 & 84.25 & 68.00 & 66.50 & 70.20 & 82.55 & 79.87  \\

\rowcolor[HTML]{FFC2C2} {RAVID (N=13)} & Openflamingo & Jpeq \(q= 60\) & 94.49 & 94.47 & 81.23 & 62.09 & 95.42 & 91.50 & 82.26 & 89.72 & 53.20 & 82.07 & 82.23 & 65.55 & 86.85 & 63.35 & 85.45 & 69.45 & 68.20 & 71.30 & 77.10 & 78.73  \\

\rowcolor[HTML]{FFC2C2} {RAVID (N=13)} & Openflamingo & Jpeq \(q= 50\) & 93.83 & 93.30 & 80.05 & 60.07 & 94.58 & 93.72 & 77.91 & 87.22 & 52.28 & 78.78 & 76.04 & 66.80 & 85.55 & 62.10 & 85.70 & 68.75 & 67.60 & 70.15 & 74.90 & 77.33  \\

\rowcolor[HTML]{FFC2C2} {RAVID (N=13)} & Openflamingo & Jpeq \(q= 40\) &  93.01 & 91.98 & 74.98 & 57.05 & 92.99 & 92.90 & 74.01 & 80.83 & 52.51 & 77.20 & 75.08 & 63.70 & 81.65 & 60.40 & 81.60 & 66.90 & 66.50 & 68.10 & 72.30 & 74.93 \\ \midrule

C2P-CLIP & - & Average & 89.68 & 91.07 & \bf84.91 & \bf69.99 & 93.57 & 91.75 & \bf82.04 & \bf91.84 & \bf53.54 & 74.96 & 74.55 & 59.62 & 17.29 & 48.52 & 18.25 & 46.24 & 47.84 & 43.76 & 25.84 & 63.44 \\

\rowcolor[HTML]{FFC2C2} {RAVID (N=13)} & Openflamingo & Average & \bf90.03 & \bf91.26 & 84.68 & 69.95 & \bf93.81 & \bf93.01 & 81.69 & 91.21 & 53.25 & \bf76.79 & \bf76.67 & \bf67.86 & \bf88.84 & \bf72.51 & \bf88.34 & \bf73.46 & \bf72.83 & \bf74.62 & \bf84.4 & \bf80.27 \\

\bottomrule
\end{tabular}
\end{adjustbox}
\end{table*}

%\multicolumn{1}{l}{RAVID (N=3) W/ RAG} & QWen-VL & Blur 2 &  &  &  &  &  &  &  &  &  &  &  &  &  &  &  &  &  &  &  &  \\

\begin{table*}[!ht]
\caption{Performance comparison of different Vision-Language Models (VLMs) on RAVID’s detection task.}
\label{tab:diffVLMs}
\begin{adjustbox}{width=\linewidth}
\begin{tabular}{@{}lrcccccccccccccccccccc@{}}
\toprule
\multicolumn{1}{c}{\multirow{2}{*}{VLMs}} & \multirow{2}{*}{$N$} & \multicolumn{6}{c}{GAN} & \multirow{2}{*}{\begin{tabular}[c]{@{}c@{}}Deep\\ Fakes\end{tabular}} & \multicolumn{2}{c}{Low level} & \multicolumn{2}{c}{Perceptual loss} & \multirow{2}{*}{Guided} & \multicolumn{3}{c}{LDM} & \multicolumn{3}{c}{Glide} & \multirow{2}{*}{Dalle} & \multirow{2}{*}{mAcc} \\ \cmidrule(lr){3-8} \cmidrule(lr){10-13} \cmidrule(lr){15-20}
\multicolumn{1}{c}{} &  & \begin{tabular}[c]{@{}c@{}}Pro-\\ GAN\end{tabular} & \begin{tabular}[c]{@{}c@{}}Cycle-\\ GAN\end{tabular} & \begin{tabular}[c]{@{}c@{}}Big-\\ GAN\end{tabular} & \begin{tabular}[c]{@{}c@{}}Style-\\ GAN\end{tabular} & \begin{tabular}[c]{@{}c@{}}Gau-\\ GAN\end{tabular} & \begin{tabular}[c]{@{}c@{}}Star-\\ GAN\end{tabular} &  & SITD & SAN & CRN & IMLE &  & \begin{tabular}[c]{@{}c@{}}200\\ steps\end{tabular} & \begin{tabular}[c]{@{}c@{}}200\\ w/cfg\end{tabular} & \begin{tabular}[c]{@{}c@{}}100\\ steps\end{tabular} & \begin{tabular}[c]{@{}c@{}}100\\ 27\end{tabular} & \begin{tabular}[c]{@{}c@{}}50\\ 27\end{tabular} & \begin{tabular}[c]{@{}c@{}}100\\ 10\end{tabular} &  &  \\ \cmidrule(r){1-2} \cmidrule(lr){9-9} \cmidrule(lr){14-14} \cmidrule(l){21-22} 
\rowcolor[HTML]{D4E8E7} Gemma3  & 1 & 77.84 & 67.68 & 61.12 & 57.39 & 77.17 & 70.79 & 55.71 & 59.17 & 61.87 & 50.10 & 50.09 & 49.20 & 70.95 & 61.80 & 70.65 & 69.95 & 69.80 & 71.00 & 60.70 & 63.84  \\

\rowcolor[HTML]{D4E8E7} Gemma3  & 3 & 92.46 & 89.82 & 76.88 & 82.34 & 90.81 & 70.94 & 59.93 & 71.94 & 63.24 & 56.83 & 56.86 & 58.90 & 86.20 & 78.10 & 85.35 & 83.55 & 83.10 & 84.40 & 79.85 &  76.19 \\

\rowcolor[HTML]{D4E8E7} Gemma3 & 5 & 96.03 & 91.56 & 86.30 & 88.60 & 94.54 & 83.62 & 68.81 & 72.50 & 64.84 & 63.61 & 63.63 & 64.15 & 91.50 & 86.95 & 91.55 & 88.65 & 89.35 & 89.90 & 88.60 &  82.35 \\

\rowcolor[HTML]{D4E8E7} Gemma3  & 7 & 97.10 & 93.79 & 92.73 & 91.06 & 96.59 & 89.64 & 69.95 & 75.83 & 64.38 & 80.74 & 80.77 & 65.70 & 94.00 & 91.10 & 93.95 & 91.00 & 91.20 & 92.05 & 92.20 & 86.52 \\

\rowcolor[HTML]{D4E8E7} Gemma3  & 13 & 97.34 & 92.73 & 92.92 & 89.68 & 95.55 & 93.42 & 76.11 & 72.22 & 62.33 & 88.62 & 88.37 & 67.25 & 95.60 & 92.45 & 95.55 & 92.50 & 92.90 & 93.30 & 93.60 & 88.02 \\ \midrule

\rowcolor[HTML]{9FC0FD}  Qwen-VL & 1  &  91.29 & 93.60 & 83.60 & 66.23 & 80.19 &  93.70 & 90.84 & 88.89 & 59.36 & 88.89 & 94.58 & 63.01 & 94.89 & 88.39 & 94.89 & 89.09 & 89.44 & 90.84 & 90.69 & 85.92 \\
 
\rowcolor[HTML]{9FC0FD} Qwen-VL  & 3 & 99.49 & 97.27 & 94.77 & 93.16 & 96.11 & 99.77 & 92.45 & 93.33 & 62.33 & 97.91 & 98.00 & 66.60 &  99.00 & 95.55 & 99.10 & 92.90 & 93.05 & 94.55 & 97.75 &  92.79 \\

\rowcolor[HTML]{9FC0FD} Qwen-VL  & 5 & 99.92 & 97.80 & 98.10 & 95.84 & 99.12 & 99.85 & 93.01 & 93.61 & 62.79 & 97.01 & 97.03 & 67.60 & 99.30 & 96.40 & 99.30 & 93.70 & 94.10 & 95.00 & 98.25 &  93.56 \\ 

\rowcolor[HTML]{9FC0FD} Qwen-VL  & 7 & 99.69 & 96.29 & 95.95 & 94.02 & 97.57 & 99.85 & 93.10 & 92.22 & 61.87 & 97.30 & 97.32 & 66.95 & 99.25 & 95.60 & 99.20 & 93.45 & 93.85 & 94.70 & 97.55 & 92.93 \\ 

\rowcolor[HTML]{9FC0FD} Qwen-VL  & 13 & 99.96 & 97.84 & 98.70 & 95.24 & 99.28 & 99.82 & 93.36 & 93.61 & 63.01 & 96.47 & 96.46 & 67.80 & 99.30 & 96.75 & 99.40 & 94.05 & 95.00 & 95.70 & 98.40 &  93.69 \\ \midrule
  
\rowcolor[HTML]{FFC2C2} Openflamingo & 1 & 50.00 & 50.00 & 50.00 & 50.00 & 50.00 & 50.00 & 50.08 & 50.00 & 50.00 & 50.00 & 50.00 & 50.00 & 50.00 & 50.00 & 50.00 & 50.00 & 50.00 & 50.00 & 50.00 & 50.00 \\

\rowcolor[HTML]{FFC2C2} Openflamingo  & 3 & 99.95 & 97.84 & 99.25 & 95.94 & 99.38 & 99.85 & 92.64 & 93.61 & 62.33 & 97.55 & 97.59 & 66.95 & 99.35 & 96.35 & 99.35 & 93.10 & 93.25 & 94.40 & 98.40 & 93.53 \\

\rowcolor[HTML]{FFC2C2} Openflamingo & 5 & 99.95 & 97.58 & 99.28 & 96.24 &  99.30 & 99.82 & 93.19 &  93.89 & 63.24 &  96.26 & 96.25 & 68.10 & 99.20 & 96.45 & 99.30 & 94.00 & 94.35 &  95.10 & 98.30 & 93.67 \\

\rowcolor[HTML]{FFC2C2} Openflamingo & 7 & 99.96 & 97.46 & 99.15 & 96.24 & 99.32 & 99.80 & 93.30 & 94.72 & 63.93 & 96.04 & 96.03 & 68.30 & 99.25 & 96.65 & 99.35 & 94.00 & 94.70 & 95.30 & 98.15 & 93.77 \\ 

\rowcolor[HTML]{FFC2C2} Openflamingo  & 13 & 99.98 & 97.35 & 99.15 & 96.27 & 99.33 & 99.82 & 93.47 & 95.00 & 63.70 & 95.53 & 95.56 & 68.75 & 99.20 & 96.95 & 99.35 & 94.65 & 94.90 & 95.85 & 98.35 & 93.85 \\

\bottomrule
\end{tabular}
\end{adjustbox}
\end{table*}

\subsection{Relevance of CLIP Task-Tuning for Contextual Image Retrieval}

To assess the impact of fine-tuning the CLIP image encoder on retrieving more relevant context, we compare two configurations: pretrained CLIP (without fine-tuning) and fine-tuned RAVID\_CLIP. In both cases, retrieved images are used as context for the Openflamingo vision-language model, with retrieval set to 1, 3, 5, 7 and 13 images.

\begin{figure*}[!th]
   \centering
   \includegraphics[width=\linewidth]{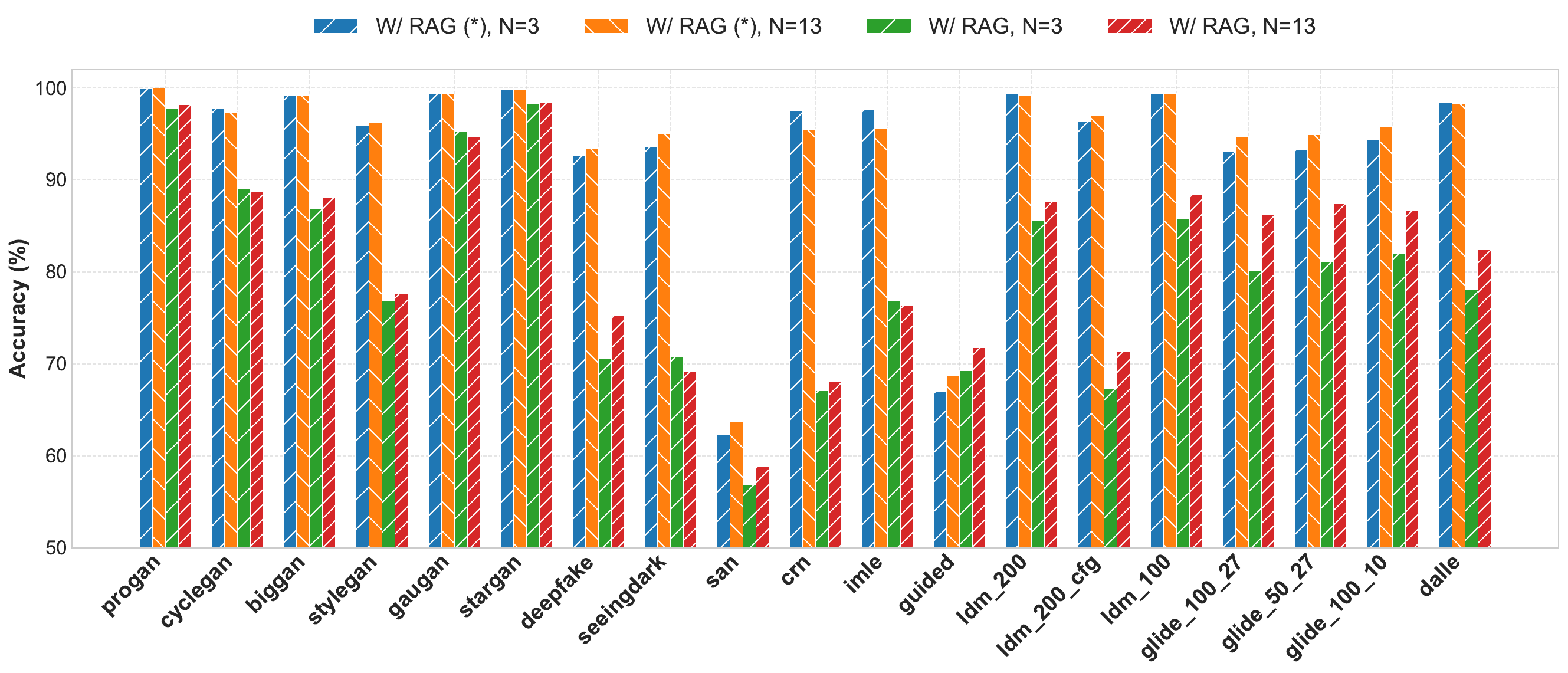}
   \caption{Evaluation of CLIP Task-Tuning's Impact on Contextual Image Retrieval. We compare image retrieval quality between the original pretrained CLIP and a fine-tuned variant (RAVID\_CLIP), evaluating how task-specific tuning affects the relevance of retrieved images in RAVID detection capacity.}
   \label{fig:tunedRelevance}
\end{figure*}

Table~\ref{tab:RAVIDICLIPnoFune} shows that when a single image is retrieved (N=1), both configurations give random results with a mAcc of 50\%. However, as the number of images retrieved increases, the benefits of fine-tuning become clear. For instance, with N=3, RAVID with the pre-trained CLIP achieves a mAcc of 79.79\%, while with the fine-tuned CLIP, it reaches 93.53\%, an absolute improvement of around 13.7\%. Similar gains occur for N=5, 7, and 13, where RAVID with fine-tuned CLIP consistently outperforms the pre-trained version by around 12 to 12.5\% in average accuracy. Improvements are not only evident at average accuracy but are also significant across multiple generative models. With N=3, notable gains can be seen in the individual model subsets: for example, StyleGAN accuracy increased from 76.91\% → 95.94\%, while BigGAN improved from 86.93\% → 99.25\%. The effect is also striking in the DeepFakes subset, where accuracy jumped from 70.53\% → 92.64\%, highlighting the ability of fine-tuned embeddings to enhance model generalization. This trend remains consistent across other retrieval settings (N=5, 7, and 13), where fine-tuned CLIP embeddings yield substantial improvements across nearly all generative models. Figure~\ref{fig:tunedRelevance} shows the impact of fine-tuning the CLIP image encoder.  The W/ RAG RAVID (*) bars (representing RAVID with fine-tuned CLIP) consistently exhibit better accuracy than the W/ RAG RAVID bars (RAVID with pre-trained CLIP) for N=3 and N=13, suggesting that fine-tuning improves the retrieval of relevant context, resulting in better detection performance.

These findings suggest that pre-trained CLIP lacks sensitivity to AI-generation artifacts, as it has been originally trained for generic vision-language tasks rather than forensic detection. Fine-tuning aligns the embedding space with AI-generated image distributions, improving retrieval effectiveness and enabling the model to extract subtle features that align with the context of AI-generated image detection. This analysis highlights the critical role of domain-specific fine-tuning in retrieval-augmented AI-generated image detection. While trained CLIP captures general visual semantics, it fails to retrieve the most relevant images for forensic analysis. In contrast, fine-tuned RAVID\_CLIP learns generative-aware representations, ensuring that retrieved images provide meaningful context, ultimately leading to state-of-the-art detection performance.

\subsection{Robustness to Image Degradation}

To systematically evaluate the robustness of our proposed approach, we assess its performance under two common forms of image degradation: Gaussian blur and JPEG compression. These perturbations simulate real-world challenges where images undergo quality loss due to compression artifacts or motion blur, which can adversely impact AI-generated image detection. We compare our method, RAVID (N=13) W/ RAG Openflamingo, against the baseline C2P-CLIP, analyzing their degradation trends across multiple generative models. The quantitative results are summarized in Table~\ref{tab:RAVIDICLIPDegradation}, and the performance trends under different blur and JPEG compression levels are illustrated in Figure~\ref{fig:deg}.

\subsubsection{Gaussian Blur Degradation}
We apply Gaussian blur with increasing standard deviations ($\sigma = 1, 2, 3$) to evaluate the resilience of both methods under varying levels of spatial smoothing. As expected, performance degrades as blur severity increases, but RAVID exhibits significantly higher robustness. At $\sigma = 1$, RAVID achieves a mean accuracy (mAcc) of 89.59\%, substantially outperforming C2P-CLIP (55.37\%). Notably, under severe degradation $\sigma = 3$, RAVID maintains an accuracy of 76.46\%, whereas C2P-CLIP deteriorates to 55.63\%. These results highlight the effectiveness of retrieval-augmented generation in preserving discriminative features even under heavy spatial distortions.

\subsubsection{JPEG Compression Robustness}
To simulate real-world compression artifacts, we apply JPEG degradation with quality factors ranging from q=80 (minimal compression) to q=40 (high compression). Both methods experience a decline in accuracy as compression severity increases, shown in Figure~\ref{fig:deg}. However, RAVID consistently outperforms C2P-CLIP across all quality levels. At q=80, RAVID achieves 83.46\%, compared to 69.32\% for C2P-CLIP. Even at the lowest quality setting (q=40), RAVID maintains a 74.93\% accuracy, outperforming C2P-CLIP (70.02\%) despite the loss of high-frequency details. The performance gap further underscores the advantages of retrieval-augmented visual-language models in handling lossy compression artifacts.

\subsubsection{Discussion}

Across both degradation types, RAVID consistently demonstrates greater robustness compared to the baseline. This can be attributed to its retrieval-augmented mechanism, which enhances contextual understanding by leveraging external image priors. By integrating semantically relevant information, RAVID is able to recover essential details that are lost due to degradation, improving the overall detection accuracy. Notably, RAVID also outperforms traditional methods in robustness, maintaining high accuracy even under image degradations such as Gaussian blur and JPEG compression. Specifically, RAVID achieves an average accuracy of 80.27\% under degradation conditions, compared to 63.44\% for the state-of-the-art model C2P-CLIP, demonstrating consistent improvements in Gaussian blur and JPEG compression scenarios. These findings reinforce the effectiveness of retrieval-augmented visual models for real-world scenarios, where image quality is often compromised due to pre-processing pipelines, social media compression, or camera limitations.

\begin{table*}[t]
\caption{Generalization performance of methods trained on 4-class ProGAN. Results show accuracy (\%) on real and synthetic data subsets, each containing 3,000 image samples.}
\label{tab:generalization}
\begin{adjustbox}{width=\linewidth}
\begin{tabular}{l|c|cc|cccccccccc|c}
%\begin{tabular}{@{}cc>{\columncolor[HTML]{CCFFCC}}c>{\columncolor[HTML]{CCFFCC}}c>{\columncolor[HTML]{feb3b1}}c>{\columncolor[HTML]{feb3b1}}c>{\columncolor[HTML]{feb3b1}}c>{\columncolor[HTML]{feb3b1}}c>{\columncolor[HTML]{feb3b1}}c>{\columncolor[HTML]{feb3b1}}c>{\columncolor[HTML]{feb3b1}}c>{\columncolor[HTML]{feb3b1}}c>{\columncolor[HTML]{feb3b1}}c>{\columncolor[HTML]{feb3b1}}c>{\columncolor[HTML]{feb3b1}}c>{\columncolor[HTML]{feb3b1}}cc@{}}
\toprule
Methods & \#params & MS COCO & Flickr & ControlNet & Dalle3 & DiffusionDB & IF & LaMA & LTE & SD2Inpaint & SDXL & SGXL & SD3 & mAcc \\ \midrule
FatFormer & 493M & 33.97 & 34.04 & 28.27 & 32.07 & 28.10 & 27.95 & 28.67 & 12.37 & 22.63 & 31.97 & 22.23 & 35.91 & 28.18\\ 
RINE & 434M & \bf 99.80 & \bf 99.90 & \bf 91.60 & 75.00 & \bf 73.00 & 77.40 & 30.90 & 98.20 & 71.90 & 22.20 & 98.50 & 08.30 & 70.56\\ 
 C2P-CLIP & 304M & 99.67 & 99.73 & 15.10 & \bf 75.57 & 27.87 & \bf 89.56 & \bf 65.43 & 00.20 & 27.90 & \bf 82.90 & 07.17 & \bf 70.46 & 55.13\\ 

\rowcolor[HTML]{FFC2C2}  RAVID (N=13)  & - &  97.83 & 99.23 & 85.80 & 68.93 & 70.70 & 60.71 & 62.97 & 99.97 & 80.37 & 62.10 & 98.80 & 58.31 & 78.81 \\
\bottomrule
\end{tabular}
\end{adjustbox}  \vspace{-3mm}
\end{table*}

%\rowcolor{olivegreen} 
%\rowcolor{lightblueAlpha}  DeeCLIP (ours) & 306M & 97.83 & 98.50 & 86.03 & 69.33 & 71.10 & 61.37 & 63.07 & \bf 99.97 & \bf 80.57 & 62.60 & \bf98.90 & 58.61 & \bf 78.99\\

% \rowcolor{lightblueAlpha}  RAVID (ours) N=3 & - & 99.80 & 99.83 & 77.43 & 18.13 & 63.73 & 7.24 & 30.97 & 99.40 & 68.00 & 12.10 & 90.77 & 21.16 & 57.38\\

% \rowcolor[HTML]{FFC2C2}  RAVID (N=13)  & - & 99.63 & 99.80 & 82.83 & 22.10 & 69.63 & 9.41 & 33.57 & 99.70 & 70.87 & 15.37 & 92.30 & 27.37 & 60.21\\

\section{Impact of Vision-Language Model Selection}

\begin{figure}[!th]
    \centering
    \includegraphics[width=\linewidth]{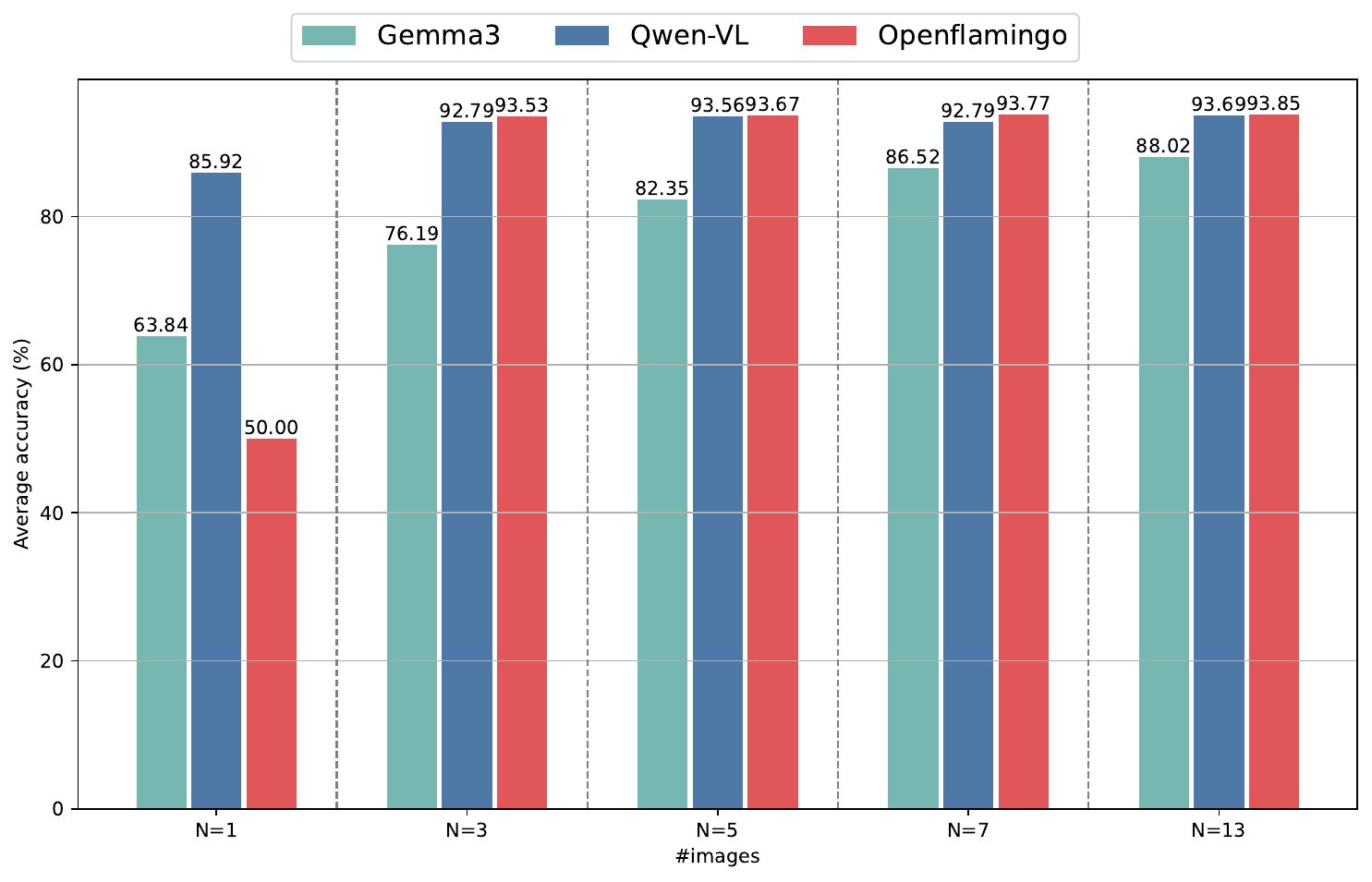}
    \caption{Impact of Vision-Language Model Selection on RAVID Performance. We assess how the vision-language models (VLMs) influence RAVID's detection capabilities by evaluating three models under varying retrieval counts (N): Qwen-VL, OpenFlamingo, and Gemma3.}
    \label{fig:VLMs}
\end{figure}

%To evaluate the influence of different vision-language models (VLMs) on RAVID’s detection ability, we conducted experiments using Qwen-VL, OpenFlamingo, and Gemma3, while varying the retrieval count (N). 
% To evaluate the influence of different vision-language models (VLMs) on RAVID’s detection ability, we conducted experiments using Qwen-VL, OpenFlamingo, and Gemma3 while varying the retrieval count (N). \hl{We initially conducted experiments using Qwen-VL in the paper. However, after additional exploration, we found that OpenFlamingo outperformed Qwen-VL in many cases, particularly with deeper retrieval counts ($N \geq 3$).} Due to computational constraints, we limited N to a maximum of 3 for Qwen-VL, while OpenFlamingo and Gemma3 were tested with higher retrieval counts. Table~\ref{tab:diffVLMs} presents the results, highlighting the varying effectiveness of each VLM.

To evaluate the influence of different vision-language models (VLMs) on the detection ability of RAVID, we conducted experiments using Qwen-VL, OpenFlamingo, and Gemma while varying the retrieval count (N). Table~\ref{tab:diffVLMs} presents the results, highlighting the varying effectiveness of each VLM. The findings provide key insights into each model's generalization ability, robustness to different generative techniques, and sensitivity to the amount of context.

\subsection*{Performance Across Generative Models}

Across generative categories, including multiple GAN variants, perceptual loss-based approaches, deepfakes, and advanced text-image diffusion models, VLMs have distinct strengths and limitations.

\textbf{Qwen-VL} consistently performs competently with its counterpart, Openflamingo, across almost all generative models. Its high detection rates, particularly in Pro-GAN, StarGAN, and diffusion-based models (e.g., LDM, Glide, and DALL·E), suggest strong multimodal understanding and generalization. Even in more challenging settings such as IMLE, and CRN, Qwen-VL maintains impressive accuracy, typically above 96\%. This robustness indicates that Qwen-VL can effectively leverage both visual and linguistic cues for fine-grained detection tasks, regardless of the manipulation’s generation mechanism.

\textbf{OpenFlamingo}, while exhibiting poor one-shot performance, a 50\% mAcc, indicative of random prediction, rapidly improves with higher shot counts. From three shots, it matches or slightly trails Qwen-VL in several categories. For example, at $N=13$, OpenFlamingo reaches a competitive 93.85\% mAcc, closely higher than Qwen-VL's 93.69\%. Its trajectory suggests a firm reliance on contextual support, potentially making it well-suited to few-shot or retrieval-augmented learning environments but less reliable in zero- or one-shot scenarios.

\textbf{Gemma3}, in contrast, exhibits a more gradual improvement as the number of shots increases. From $N=1$, we observe ongoing gains reaching 88.02\% at $N=13$. While Gemma3 performs solidly across GAN-based generations and maintains decent results on LDM and Glide models, it shows noticeable weaknesses in perceptual loss and lower-level models. This pattern suggests Gemma3 may have less sensitivity to nuanced texture or distributional artifacts that other models pick up on more easily.

\subsection*{Sensitivity to Number of Shot}

Figure~\ref{fig:VLMs} highlights the progression of average accuracy (mAcc) of VLMs with increasing retrieval count N. The trends are revealing: 

\begin{itemize}
    \item Qwen-VL shows high performance from the outset, starting at 85.92\% with just one shot and reaching near-maximum capacity, 92.79\%, after three shots. This suggests the model is inherently robust and benefits modestly from additional contexts.
    \item OpenFlamingo shows a 50\% jump to over 93\% mAcc between one and three shots. From there, performance stagnates, with minor gains up to 93.85\% at 13 shots. This pattern reflects a steep learning curve, highlighting its adaptability in few-shot contexts.
    \item Gemma3 improves more linearly, its mAcc rising consistently from 63.84\% at $N=1$ to 88.02\% at $N=13$. Unlike OpenFlamingo, it shows a slower rate of improvement but a consistent trajectory, which might favor use cases where the retrieval count is expected to scale over time.
\end{itemize}

These trends indicate a practical trade-off between initial generalization capacity and few-shot adaptability. Qwen-VL is a strong candidate for out-of-the-box performance or low-shot applications, while OpenFlamingo is preferable in scenarios with a richer support context available. Gemma3’s progressive improvement makes it suitable for pipelines where inference is augmented incrementally.

\subsection*{Implications for RAVID}

These results underscore the value of advanced vision-language alignment in improving AI-generated image detection. Qwen-VL’s superior performance suggests that high-capacity models with strong vision-language fusion mechanisms can reliably detect images from various generative models. Furthermore, the stark contrast between OpenFlamingo’s one-shot and few-shot performance reveals the importance of retrieval augmentation in practical deployments of VLMs on AI-generated image detection tasks. In practical settings, the choice of model and retrieval strategy should be guided by task constraints: Qwen-VL excels in a few context situations, OpenFlamingo benefits from enriched support, and Gemma3 provides a consistent baseline.

% Among the tested vision language models (VLMs), OpenFlamingo achieved the highest accuracy (93.85\%) at N=13, followed closely by Qwen-VL (92.79\%) at N=3. In contrast, Gemma3 underperformed relative to the other models, peaking at 82.35\% at N=5.

% \begin{itemize}
%     \item \textbf{Qwen-VL:} Demonstrates strong performance, achieving 85.92\% accuracy at N=1 and improving significantly to 92.79\% at N=3. However, we were unable to test N $>$ 3 due to computational constraints, leaving open the question of whether additional retrieved images would further enhance performance.

%     \item \textbf{OpenFlamingo:} Unlike Qwen-VL, OpenFlamingo performed poorly at N=1 (50.00\%), indicating a strong dependency on retrieval for meaningful predictions. However, its accuracy increased drastically to 93.53\% at N=3 and peaked at 93.85\% at N=13.

%     \item \textbf{Gemma 3:} Achieves 76.19\% accuracy at N=3 and 82.35\% at N=5, showing a more gradual performance improvement but remaining less effective than the other VLMs.
% \end{itemize}

% \noindent In summary, Qwen-VL and OpenFlamingo outperform Gemma3 in retrieval-augmented detection, with Qwen-VL excelling even with limited retrieval counts while OpenFlamingo requires higher retrieval counts, highlighting that VLM selection impacts detection performance beyond just the number of retrieved images and suggesting future exploration of hybrid approaches for improved generalization.

\section{Generalization on Unseen Data}

To assess cross-domain robustness, we evaluate the generalization performance of four top-performing methods from Table~\ref{tab:TrainedOnUniversalFakeDetect}, FatFormer, RINE, C2P-CLIP, and RAVID. Each model is trained solely on the ProGAN 4-class dataset and tested on a broad spectrum of unseen data sources, including authentic images (e.g., MS COCO, Flickr) and diverse generative models (e.g., ControlNet, DALL·E 3, DiffusionDB, SGXL, LTE, etc...). This evaluation simulates a realistic open-world scenario where detection models face data distributions that deviate significantly from their training domain.

As shown in Table~\ref{tab:generalization}, RAVID exhibits strong cross-domain generalization, achieving the highest overall mean accuracy of 78.81\% across both real and synthetic domains. While other methods demonstrate strengths in specific datasets, they often suffer from instability under distribution shifts. RINE, for instance, performs exceptionally well on MS COCO and Flickr but collapses on SD3, reflecting limited robustness to unseen generative processes. C2P-CLIP attains high accuracy on IF and DALL·E 3 but fails drastically on LTE, revealing poor transferability under localized editing. In contrast, RAVID maintains consistent and competitive performance across nearly all test domains, particularly excelling on challenging sources like LTE (99.97\%), SGXL (98.80\%), and SD2Inpaint (80.37\%). Despite being trained on a single synthetic domain, its strong performance on real and synthetic datasets underscores its capacity to learn transferable representations, making it a compelling candidate for reliable deployment in open-world scenarios.
\section{Conclusion}
\label{sec:conclusion}
In this paper, we have introduced RAVID, a novel retrieval-augmented framework for detecting AI-generated images. By dynamically retrieving and integrating relevant visual knowledge, RAVID enhances detection accuracy and generalization. Unlike traditional methods that rely on low-level artifacts or model fingerprints, our approach leverages a fine-tuned CLIP-based image encoder (RAVID\_CLIP) for embedding generation and retrieval. Additionally, we incorporate \acp{vlm} like Openflamingo to enrich contextual understanding. Retrieving semantically relevant images from a vector database and integrating them into the detection pipeline significantly improves performance. Evaluations on the UniversalFakeDetect benchmark (spanning 19 generative models) showed that RAVID outperforms existing methods, achieving 93.85\% accuracy in both in- and out-of-domain settings. A detailed analysis revealed a 35.51\% performance gap between setups W/ and W/O retrieval in the 3-shot setting, highlighting the critical role of relevant context. Our findings confirm that retrieval not only enhances accuracy but also scales with additional retrieval shots, reinforcing its impact. Furthermore, our robustness analysis demonstrated that RAVID maintains superior detection performance under commun image degradations, including Gaussian blur and JPEG compression. In contrast baseline methods struggle with spatial distortions and compression artifacts, RAVID leverages retrieval-augmented generation to incorporate relevant context, maintaining robust performance and enabling significantly higher accuracy across all tested degradation levels. It achieves an average accuracy of 80.27\% under degradation conditions, compared to 63.44\% for the state-of-the-art model C2P-CLIP, demonstrating consistent improvements in both Gaussian blur and JPEG compression scenarios. These results highlight the resilience of retrieval-based approaches in real-world conditions where images often suffer from quality loss due to compression pipelines, motion blur, or other distortions. As generative models rapidly evolve, retrieval-augmented techniques like RAVID will be essential for developing more robust, adaptable AI-generated content detection systems. Beyond improving detection accuracy, RAVID paves the way for advancements in context-aware and resilient detection mechanisms.

%\hl{better to talsk about delta W:O RAG to RAG}. 

\section*{Acknowledgments} This work has been partially funded by the project PCI2022-134990-2 (MARTINI) of the CHISTERA IV Cofund 2021 program. Abdenour Hadid is funded by TotalEnergies collaboration agreement with Sorbonne University Abu Dhabi.

%\newpage

{
    \small
    \bibliographystyle{ieeenat_fullname}
    \bibliography{main}
}

\end{document}